\documentclass[a4paper,fleqn]{cas-sc}

\usepackage[numbers]{natbib}

\usepackage{amsmath,bm}          
\usepackage[protrusion=true,expansion=false]{microtype}

\usepackage{booktabs,longtable,array,tabularx,multirow}

\usepackage{graphicx}
\usepackage{subcaption}
\usepackage{float}
\usepackage{placeins}

\usepackage{cleveref}

\usepackage{enumitem}

\usepackage{xcolor}
\usepackage{tcolorbox}
\tcbuselibrary{skins, breakable}

\usepackage{tikz}
\usetikzlibrary{shapes.geometric, arrows.meta, positioning,
                fit, backgrounds, shadows, calc}
\usepackage{pgfplots}
\pgfplotsset{compat=1.18}
\usepgfplotslibrary{groupplots}

\usepackage{mdframed}
\usepackage{url}
\usepackage{lastpage}   
\usepackage{fancyhdr}  
\usepackage{pifont}    

\definecolor{darkblue}{RGB}{18,52,86}
\definecolor{tealc}{RGB}{0,128,96}
\definecolor{purplec}{RGB}{110,40,160}
\definecolor{amberc}{RGB}{180,110,0}
\definecolor{redc}{RGB}{160,30,20}
\definecolor{plgray}{RGB}{245,245,245}
\definecolor{plborder}{RGB}{0,84,166}
\definecolor{s1col}{RGB}{0,84,166}
\definecolor{s2col}{RGB}{0,128,96}
\definecolor{s3col}{RGB}{160,50,0}

\hypersetup{colorlinks=true, linkcolor=darkblue,
            citecolor=tealc, urlcolor=darkblue,
            pdftitle={SegTME-UNI2 Paper 1}}

\newcommand{\fw}{\text{SegTME-UNI2}}
\newcommand{\seg}{\text{UNI2-UperHoVer}}
\newcommand{\bb}{\text{UNI2-h}}
\newcommand{\eg}{\textit{e.g.}}
\newcommand{\ie}{\textit{i.e.}}

\newcommand{\loss}[1]{\mathcal{L}_{\text{#1}}}

\tikzset{
  pbox/.style={rectangle, rounded corners=4pt, minimum width=#1,
               minimum height=1.0cm, align=center, draw,
               font=\footnotesize\bfseries, inner sep=5pt},
  pbox/.default=2.6cm,
  sbox/.style={rectangle, rounded corners=3pt, minimum width=#1,
               minimum height=0.78cm, align=center, draw,
               font=\scriptsize, inner sep=4pt},
  sbox/.default=2.2cm,
  arr/.style={-{Stealth[length=4pt]}, thick},
  darr/.style={-{Stealth[length=4pt]}, thick, dashed},
  grp/.style={rectangle, rounded corners=6pt, draw, dashed, inner sep=8pt},
  lbl/.style={font=\scriptsize\bfseries, align=center},
  slbl/.style={font=\tiny, align=center},
  s1B/.style={fill=s1col!20, draw=s1col},
  s1A/.style={fill=s1col!10, draw=s1col!60},
  s2B/.style={fill=s2col!20, draw=s2col},
  s2A/.style={fill=s2col!10, draw=s2col!60},
  s3B/.style={fill=s3col!20, draw=s3col},
  s3A/.style={fill=s3col!10, draw=s3col!60},
  colFPN/.style={fill=amberc!15, draw=amberc!70},
  colSem/.style={fill=s1col!12, draw=s1col!60},
  colHV/.style={fill=purplec!12, draw=purplec!60},
  colBio/.style={fill=tealc!12, draw=tealc!60},
  colNV/.style={fill=tealc!20, draw=tealc},
}

\begin{document}
\let\WriteBookmarks\relax
\renewcommand{\floatpagefraction}{0.1}
\renewcommand{\textfraction}{0.1}
\renewcommand{\topfraction}{0.9}
\renewcommand{\bottomfraction}{0.9}

\pagestyle{fancy}
\fancyhf{}
\renewcommand{\headrulewidth}{0pt}
\fancyfoot[C]{\thepage\ of \pageref{LastPage}}

\shorttitle{SegTME-UNI2: Foundation Model for TME Characterisation}
\shortauthors{W.S.H.M.W.~Ahmad et~al.}

\title[mode=title]{SegTME-UNI2: A Foundation Model-Based Framework for
Generalisable Multiclass Cell Segmentation and
LLM-Driven Tumour Microenvironment Characterisation
in Histopathology}

\author[1]{Wan Siti Halimatul Munirah Wan Ahmad}[orcid=0000-0001-6364-6341]
\cormark[1]
\ead{munirahy@sunway.edu.my}
\credit{Conceptualization, Methodology, Software, Writing (original draft), Writing (review \& editing)}

\author[1,2]{Faris Syahmi Samidi}
\credit{Software, Resources}

\author[1]{Mohammad Badal Ahmmed}
\credit{Formal analysis}

\author[1]{Vimal Angela Thiviyanathan}
\credit{Data curation}

\author[3]{Selvam Thavaraj}
\credit{Validation, Writing (review \& editing)}

\author[1]{Anwar P.P. Abdul Majeed}
\credit{Methodology, Writing (review \& editing)}

\affiliation[1]{organization={Department of Data Science and Artificial Intelligence,
  School of Computing and Artificial Intelligence,
  Faculty of Engineering and Technology, Sunway University},
  city={Bandar Sunway},
  country={Malaysia}}

\affiliation[2]{organization={Department of Electrical and Electronic Engineering, College of Engineering, Universiti Tenaga Nasional},
  city={Kajang},
  country={Malaysia}}

\affiliation[3]{organization={Faculty of Dentistry, Universiti Malaya},
  city={Kuala Lumpur},
  country={Malaysia}}

\cortext[1]{Corresponding author}

\begin{abstract}
Characterising the tumour microenvironment (TME) from routine
haematoxylin and eosin (H\&E)-stained histology images requires
simultaneous cell segmentation, biological feature extraction, and
interpretable clinical reporting. We present \fw{}, a unified
framework addressing all three requirements end to end: a segmentation
backbone that converts raw H\&E patches into per-nucleus class labels,
a structured feature-extraction pipeline that turns those labels into
quantitative TME descriptors, and a language-model narrative generator
that turns those descriptors into clinician-readable text.
At its core is \seg{}, a dual-head multiscale segmentation model that pairs
the \bb{} pathology foundation model (ViT-Giant, pretrained on over
200 million tiles from 350,000+ slides) with two parallel UperNet
decoders: one for six-class semantic segmentation and one for
horizontal-vertical (HV) gradient regression enabling
watershed-based nuclear instance separation. It is trained via a three-stage
progressive pseudo-label curriculum, scaling from
7,901 human-annotated PanNuke patches (Stage~1, 0.25~$\mu$m/pixel)
to 271,711 TCGA-UT Scale-0 patches (Stage~2, 0.5~$\mu$m/pixel) and the
full 1.6M-patch, six-scale TCGA-UT corpus (Stage~3, 0.5 to
1.0~$\mu$m/pixel), with no weights transferred between stages. TCGA-UT's
coarser, broader per-patch context than PanNuke's also permits a larger
inter-tile stride during whole-slide inference, covering the same tissue area
with substantially fewer tiles and making the adapted backbone more efficient
for large-tile and whole-slide inference. 
Segmentation operates over large, variable-size tissue tiles, on the order of
1--2\,mm, in comparison of the usual small fixed-size patches, giving the downstream TME
feature-extraction pipeline enough spatial context to characterise each tile
accurately. This pipeline
computes 22 per-patch compositional, morphological, spatial-entropy,
and intercellular-distance metrics and translates them into six categorical
phenotype labels and a standardised biological-token vocabulary, which
a LLaMA-3.2-1B model fine-tuned
via NVIDIA BioNeMo (\texttt{llama3\_native\_te}) converts into clinically
grounded narratives whose individual claims can be spot-checked directly
against the underlying numeric features, reducing
hallucination risk relative to image-conditioned generation. Given prior tiling of a
whole-slide image, the framework produces one such narrative per large tile. The
prompt format reserves metadata fields for indexing and aggregating these tile-level
narratives into a slide-level description, an extension the framework is designed to
support.
Qualitative
validation on five multi-centric, multi-scanner IGNITE NSCLC tiles, reviewed by a
single pathologist as a proof-of-concept check, shows the
pipeline produces biologically coherent phenotype classifications and
narratives despite inter-institutional stain variability and imperfect
segmentation. The robustness is attributable to grounding narratives in aggregate
population-level statistics, and not focusing on single-cell boundary precision.
The pseudo-labelled TCGA-UT dataset and \seg{}
checkpoints are publicly released to support large-scale TME
profiling and spatial biology research.
\end{abstract}

\begin{highlights}
\item UNI2-UperHoVer: dual-head, multiscale (UperNet) nuclear instance separation
\item Unsupervised curriculum bridges 200-fold PanNuke-to-TCGA-UT gap (0.25 to 0.5~$\mu$m/px)
\item LargeTilePredictor (~1-2mm tiles) enables efficient whole-slide inference
\item TME extraction turns segmentation into grounded tokens for characterisation
\item BioNeMo turns TME tokens into short, verifiable clinical narratives
\end{highlights}


\maketitle

\section{Introduction}

Histopathological examination of haematoxylin and eosin (H\&E)-stained tissue sections
is the clinical gold standard for cancer diagnosis, staging, and treatment planning.
Making sense of these images computationally means solving two related problems:
(i)~\textit{semantic segmentation}, labelling each pixel by cell type (tumour, immune,
stroma, and so on), and (ii)~\textit{nuclear instance separation}, telling individual
cells apart so they can be counted and measured. Together, these let us quantify the
tumour microenvironment (TME): how much of the tissue is tumour, how many immune cells
have infiltrated it, and how close those immune cells sit to the tumour, all of which
carry prognostic value across many cancer types~\cite{saltz2018spatial,kather2019deep}.

Existing tools fall short of doing this well at scale. Models such as
HoVer-Net~\cite{graham2019hover} can tell individual cells apart, but they need training
data where every nucleus has been manually outlined, which only exists for small
benchmark datasets such as PanNuke~\cite{gamper2020pannuke} (7,901 patches, 189,744
annotated nuclei) and CoNSeP~\cite{graham2019hover}. These datasets are also captured
at a very fine resolution (0.25\,$\mu$m/pixel), so each 256$\times$256-pixel patch
covers only 64$\times$64\,$\mu$m of tissue, roughly a few dozen cells across. That is
enough detail to outline a nucleus accurately, but not enough surrounding context to
judge broader tissue patterns, such as how immune cells are distributed relative to the
tumour. It also has a practical cost: to scan a large region of tissue at this
resolution, a model has to step across it in very small increments, so many more
patches must be processed to cover the same area. Semantic segmentation models such as
UperNet~\cite{xiao2018unified}, which pool features at multiple scales, handle broader
context and coarser resolutions better, but cannot separate touching nuclei; neither
type of model on its own produces output a pathologist could read directly as a report.

Large public repositories such as The Cancer Genome Atlas (TCGA) offer a way around the
annotation bottleneck. TCGA hosts digitised whole-slide images from more than 30,000
patients across 33 cancer types, but almost none of it has pixel-level cell
annotations, since manually outlining every nucleus in millions of patches is not
feasible. The TCGA-UT Restructured subset used in this work covers 32 solid cancer
types from 7,175 patients across 8,736 slides, comprising 1.6M~patches without any
human pixel-level labels. Beyond its scale, TCGA-UT is also captured at a coarser
resolution than PanNuke, so each patch covers substantially more physical tissue, from
$128\times128\,\mu$m up to $256\times256\,\mu$m depending on scale
(\Cref{sec:tcga_dataset}). This matters practically, not just statistically. When
scanning a slide, the model steps across it in increments called a stride, and that
stride can be as large as the tile itself. A model working at this coarser resolution
can therefore take larger steps and cover the same slide with far fewer tiles, whereas
a model working at PanNuke's fine resolution would need much smaller steps, and many
more of them, to cover the same ground. \fw{}'s
large-tile predictor (\Cref{sec:inference}) is built to exploit this directly: it accepts
variable-size tiles and scales its stride accordingly. Closing the roughly 200-fold gap between
PanNuke's labelled patches and TCGA-UT's unlabelled ones, while also taking advantage of
TCGA-UT's broader context per patch, is the central challenge \fw{} addresses.

Large-scale pathology foundation models have recently made this gap easier to bridge.
UNI2~\cite{chen2024uni2}, pretrained on over 200 million tiles from 350,000+ slides
across 40+ cancer types, produces general-purpose tile representations that transfer
well across cancer types and scanners. Its architecture also allows dense-prediction
decoder heads to be attached directly, so it can be adapted for pixel-level
segmentation without retraining the whole backbone. We pair it with UperNet, whose
multi-scale pooling suits a model that must work across both PanNuke's fine resolution
and TCGA-UT's coarser ones.

The remaining gap is turning these pixel-level outputs into something a pathologist can
actually use. Numbers like cell counts, spatial entropy, or immune-to-tumour distance
are hard to interpret without narrative context, and clinicians reasonably expect some
explanation for whatever a model concludes. We use NVIDIA
BioNeMo~\cite{nvidia2023bionemo} to fine-tune \texttt{meta-llama/Llama-3.2-1B} (via its
\texttt{llama3\_native\_te} recipe) so that it turns structured, quantitative TME
features into short, clinically-styled narratives. Prior work has tackled segmentation,
foundation models, or language-based reporting individually, but rarely combines all
three into one framework: existing approaches that come close either rely on small
manually annotated datasets, stop at segmentation without biological interpretation, or
generate narrative text without grounding it in measurable features.

We present \fw{} (\textbf{Seg}mentation-based \textbf{TME} characterisation with
\textbf{UNI2}), an end-to-end framework built from three parts, all operating on large,
variable-size tissue tiles: (1)~a segmentation
model, \seg{}, that labels cell types and separates individual nuclei; (2)~a
feature-extraction pipeline that turns those labels into quantitative, patch-level TME
statistics; and (3)~a narrative generator, fine-tuned via BioNeMo, that turns those
statistics into short clinical-style text. To train the segmentation model without
instance-level annotations at scale, we use a three-stage training process: it first
learns from PanNuke's fully human-annotated data (Stage~1), then adapts to TCGA-UT's
coarser, broader-context patches using its own predictions as pseudo-labels
(Stage~2), and finally extends to the full six-scale TCGA-UT corpus the same way
(Stage~3). Each stage trains a fresh model (not continuing from the previous
one), so that each stage's starting point reflects only the quality of the pseudo-labels
it was given, not carried-over weights.

Our contributions are:

\begin{enumerate}[leftmargin=*, noitemsep, topsep=3pt]
  \item \textbf{\seg{}}: a dual-head multiscale segmentation model that adds two decoder heads,
        one for cell type and one for separating touching nuclei, onto \bb{}, trained
        so that nuclear separation can be learned from cell-type labels alone, without
        needing instance-level annotations at any stage.
  \item \textbf{A three-stage pseudo-label training process that trades resolution for
        efficient large-tile inference}: scaling from PanNuke's native resolution (0.25~$\mu$m/pixel) to
        TCGA-UT's coarser per-patch resolution (0.5~$\mu$m to 1.0~$\mu$m/pixel, with 271,711 patches at Stage~2, 1.6M~patches across six scales at Stage~3). The point of this adaptation, beyond
        enlarging the training set, is that a model trained at TCGA-UT's coarser
        resolution supports the larger inter-tile stride described above, making
        inference over large tiles and whole slides substantially more efficient than
        a model trained only at PanNuke's native resolution would allow.
  \item \textbf{A large-tile predictor for whole-slide-scale inference}: \seg{} and the
        downstream TME pipeline operate on variable-size tissue tiles on the order of
        1--2\,mm, an order of magnitude broader than the small, fixed-size patches
        typical of nucleus-segmentation benchmarks. This gives both the spatial context
        TME characterisation requires and a larger inter-tile stride for faster
        whole-slide inference, and is exposed as a standalone, installable component
        (\Cref{sec:inference}).
  \item \textbf{Structured TME extraction for grounded token description}: a pipeline
        that turns segmentation output into 22 quantitative features per patch
        (composition, morphology, spatial entropy, intercellular distance) and
        translates them into a fixed vocabulary of biological tokens via rule-based
        thresholds, producing an auditable, quantitatively grounded description of the
        tissue before any language model is involved.
  \item \textbf{BioNeMo narrative generation for clinician-facing interpretation}: a
        LLaMA-3.2-1B model, fine-tuned via BioNeMo, converts these structured tokens
        into short, clinically-styled narrative text whose claims are traceable back
        to specific features, intended to be directly interpretable by a pathologist
        or clinician).
\end{enumerate}

\section{Related Work}

\subsection{Nuclear instance segmentation}
HoVer-Net~\cite{graham2019hover} simultaneously segments and classifies nuclei by
learning horizontal-vertical distance gradient maps, whose energy surface drives
watershed-based instance separation. It remains the most directly comparable baseline
to \seg{}, requiring instance annotations and uses a modified Preact-ResNet-50 encoder, whereas \seg{} uses a pathology-pretrained UNI2-h ViT-Giant backbone with broader pretraining exposure. CellPose~\cite{stringer2021cellpose} learns a flow field for cell boundaries and StarDist~\cite{schmidt2018cell} predicts star-convex polygons, where both
are effective for compact, homogeneous nuclei but handle multi-class tissue contexts less
well. CellViT~\cite{horst2024cellvit} applies a pre-trained ViT encoder to cell
segmentation via a U-Net-style decoder, demonstrating that ViT-based encoders
outperform CNN-based encoders for nucleus delineation on PanNuke and MoNuSeg.
\seg{} extends this direction by pairing a larger-scale pathology-pretrained ViT-Giant
encoder (\bb{}) with a dual-head UperNet decoder that jointly optimises semantic and
instance objectives, and by removing the dependency on instance-level training
annotations through dynamic HV target synthesis from semantic labels alone.

\subsection{Semantic tissue segmentation and decoder selection}
UperNet~\cite{xiao2018unified} combines a Pyramid Pooling Module (PPM) with an FPN for
multi-scale context aggregation, achieving strong results on ADE20K and diverse dense
prediction benchmarks. UNet and SegFormer have been applied to histopathology
segmentation~\cite{ronneberger2015unet,xie2021segformer} but could not fuse features across large spatial scale differences, such as for training
data spans 0.25 to 1.0\,$\mu$m/pixel.

Panoptic segmentation methods~\cite{kirillov2019panoptic} unify semantic and instance
prediction in a single output head but require panoptic-quality annotations
(both semantic classes and instance IDs for every pixel).
\seg{} adopts a dual-head design instead: the semantic head and HV regression head
have fundamentally different output types (discrete class probabilities
vs.\ continuous gradient fields), and the dynamic HV synthesis approach explicitly
requires separate semantic labels as input, making the dual-head factorisation an
architecture design necessity.
Furthermore, the pseudo-label curriculum operates exclusively on semantic labels,
which are substantially cheaper to generate at scale than full panoptic annotations.

\subsection{Pathology foundation models}
UNI~\cite{chen2024uni} and UNI2~\cite{chen2024uni2} demonstrated that self-supervised
pretraining on large-scale pathology slide corpora produces representations that transfer
across cancer type, staining protocol, and scanner vendor. UNI2 is pretrained on over 200
million tiles from 350,000+ slides across 40+ cancer types. CONCH~\cite{lu2023towards}
adds vision-language pretraining, aligning image and text representations via a
contrastive objective; ProvGigaPath~\cite{xu2024provgigapath} scales to gigapixel-level
WSI representations by summarising a whole slide into a single embedding. Neither
objective is designed to produce the dense, per-pixel predictions a segmentation decoder
needs. UNI2-h (ViT-Giant), by contrast, retains a spatial grid of per-tile features that
dense-prediction decoder heads can attach to directly, which is why it is chosen as the backbone for  this framework.

\subsection{Semi-supervised learning and pseudo-labelling}
Semi-supervised self-training with pseudo-labels has been applied to medical image
segmentation~\cite{bai2017semi,tarvainen2017mean}, and entropy-based confidence filtering is commonly used to reduce pseudo-label noise by retaining only high-confidence predictions~\cite{grandvalet2004semi}. In computational pathology, related strategies have been explored across nuclei segmentation, tissue semantic segmentation, and domain adaptation. CAPL-Net~\cite{Li2022} combines category-aware feature alignment with prototype-based pseudo-labelling for cross-domain nuclei instance segmentation and classification. Multi-Layer Pseudo-Supervision generates CAM-based pseudo masks from patch-level classification labels to perform histopathology tissue semantic segmentation, substantially reducing reliance on dense pixel-level annotation~\cite{Han2022}. SegGini constructs a tissue-graph representation and performs weakly-supervised segmentation via node classification, exploiting both local and global inter-tissue-region relations to learn from inexact and incomplete labels~\cite{Anklin2021}. These studies demonstrate the relevance of pseudo-labelling and weak supervision in computational pathology, but each typically targets a single task, relies on a single adaptation step, or depends on weak supervision from patch- or image-level labels. In contrast, our curriculum progressively expands from human-annotated PanNuke to TCGA-UT scale-0 patches and then to the full multi-resolution TCGA-UT corpus, using staged entropy-filtered pseudo-labelling to support multiclass cell segmentation and downstream TME characterisation.


\subsection{LLM integration in pathology}
\label{sec:llm_related_work}

Recent work has introduced vision-language foundation models for computational pathology, including PLIP~\cite{huang2023visual} and CONCH~\cite{lu2023towards}, which support image-text alignment, classification, retrieval, captioning, and other downstream pathology tasks. More directly related to generative pathology assistance, PathChat~\cite{lu2024pathchat} demonstrates a multimodal generative AI copilot that combines a pathology vision encoder with a large language model for diagnostic question answering and open-ended pathology interpretation. \fw{}, by contrast, is scoped specifically to describing TME and tumour-immune-microenvironment behaviour, limited to features related to tumour, stromal, and immune cell populations, whereas PathChat is generalised across a broader range of diagnostic tasks. Different from image-conditioned systems that generate responses directly from visual inputs, \fw{} takes a complementary approach: its narratives are grounded in verifiable quantitative TME features, including cell counts, spatial entropy, and intercellular distances. This structured feature grounding reduces hallucination risk and enables spot-checking of generated descriptions against the underlying numerical measurements. BioNeMo~\cite{nvidia2023bionemo} provides an SFT framework for domain-specific language-model adaptation, enabling controlled narrative generation from structured TME feature inputs.

\section{Methods}
\label{sec:methods}

\subsection{Datasets}

\subsubsection{PanNuke: supervised seed (Stage 1)}
PanNuke~\cite{gamper2020pannuke} comprises 7,901 H\&E image patches of size
$256\times256$ pixels, captured at $\times40$ magnification (0.25\,$\mu$m/pixel),
sourced from 19 tissue types. The dataset provides pixel-level instance segmentation
masks for 189,744 labeled nuclei across five foreground cell categories plus Background:
Neoplastic (class 1), Inflammatory (class 2), Connective (class 3), Dead (class 4),
and Non-neoplastic Epithelial (class 5). PanNuke is the \emph{only} source of
human-annotated pixel-level labels used in \fw{} and serves exclusively for Stage~1
supervised training of $\mathcal{M}_1$.

\subsubsection{TCGA-UT Restructured: large-scale unlabelled domain (Stages 2 \& 3)}
\label{sec:tcga_dataset}
The TCGA-UT dataset~\cite{komura2022universal} is a large-scale collection of
H\&E-stained histopathological image patches derived from The Cancer Genome Atlas (TCGA).
It was constructed from 32 solid cancer types, sourcing 9,662 diagnostic slides from
7,951 patients; after quality control (removal of 926 slides with poor staining, low
resolution, out-of-focus regions, or absent cancerous tissue), the final corpus comprises
8,736 slides from 7,175 patients. For each slide, trained pathologists selected
representative tumour regions as polygon annotations, from which 10 patches per region
were randomly cropped at the native scanning resolution of $0.5\,\mu\text{m/pixel}$. At
Scale~0, this native crop is delivered directly as a $256\times256$-pixel patch covering
$128\times128\,\mu\text{m}$ physical area.

\begin{figure}[htbp]
\centering
\begin{tabular}{cccccc}
\includegraphics[width=0.14\textwidth]{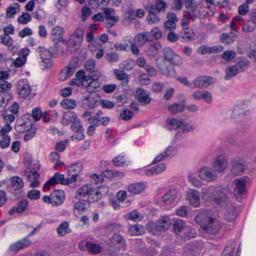} &
\includegraphics[width=0.14\textwidth]{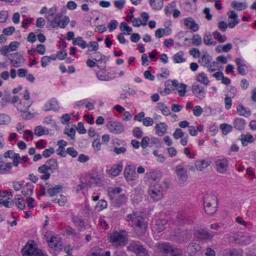} &
\includegraphics[width=0.14\textwidth]{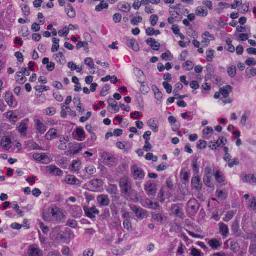} &
\includegraphics[width=0.14\textwidth]{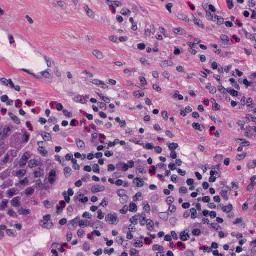} &
\includegraphics[width=0.14\textwidth]{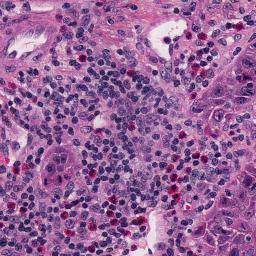} &
\includegraphics[width=0.14\textwidth]{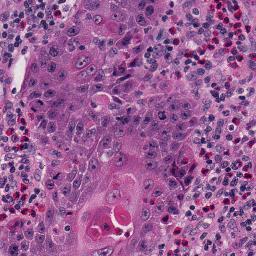} \\
Scale 0 & Scale 1 & Scale 2 &
Scale 3 & Scale 4 & Scale 5 
\end{tabular}
\caption{\textbf{TCGA-UT multi-resolution patches from the same tissue region (TCGA-BA-5558).}
Each column is delivered as the same $256\times256$-pixel file, but the physical field of
view widens from $128\times128\,\mu$m at Scale~0 (0.5\,$\mu$m/pixel, finest detail) to
$256\times256\,\mu$m at Scale~5 (1.0\,$\mu$m/pixel, broadest tissue context), since
coarser scales are constructed by cropping a progressively larger native-resolution
region and downsampling it to the same $256\times256$-pixel size (\Cref{sec:tcga_dataset}).
At finer scales, individual nuclear
morphology and staining intensity are clearly resolved; at coarser scales, intercellular
spatial organisation and tissue-level architecture become more prominent. This
multi-resolution representation enables \fw{} to learn progressively richer
contextual features across the three curriculum stages.}
\label{fig:tcga_scales}
\end{figure}

The dataset provides patches at six resolution levels (scales 0--5), with each scale
representing a progressively coarser sampling of the same underlying tissue: Scale~0 at
0.5\,$\mu$m/pixel through Scale~5 at 1.0\,$\mu$m/pixel (\Cref{tab:tcga}). This is not
documented in the original dataset release, but our own inspection (cross-checking
nuclear size, e.g.\ lymphocyte diameter, across scales) confirms that Scales~1--5 are
constructed by cropping a progressively larger native-resolution region, up to
$512\times512$ pixels at $0.5\,\mu\text{m/pixel}$ for Scale~5 (covering
$256\times256\,\mu\text{m}$ of physical tissue), and downsampling it to the same
standard $256\times256$-pixel delivery size (not by re-scanning at a coarser
native resolution). Consequently, physical tissue coverage increases from
$128\times128\,\mu\text{m}$ at Scale~0 to $256\times256\,\mu\text{m}$ at Scale~5, even
though every delivered patch file is $256\times256$ pixels.
\Cref{fig:tcga_scales} illustrates the same tissue region at all six resolutions, showing
how fine-grained nuclear details visible at Scale~0 transition to broader
tissue-architecture views at coarser scales as the effective field of view widens.

We restructured the dataset from the original per-image subdirectory layout (each image
directory containing multiple individual patch files) to a flattened per-patch layout
(each patch stored directly under its cancer-type directory), which substantially reduces
filesystem overhead during large-scale training. The restructured
dataset~\cite{tcgaUT2024} is publicly released on HuggingFace. \Cref{tab:tcga}
summarises the patch counts per scale.

\begin{table}[htbp]
\centering
\caption{TCGA-UT Restructured dataset patch counts per resolution scale.
The full corpus of 1.6M~patches spans 32 solid cancer types from TCGA
(8,736 slides, 7,175 patients; original source: Komura \& Ishikawa \cite{komura2022universal}).
All pixel-level labels for Stages 2 and 3 are model-generated pseudo-labels;
no human annotations exist for this dataset.}
\label{tab:tcga}
\small
\begin{tabular}{ccccc}
\toprule
Scale & Patches & Resolution ($\mu$m/pixel) & Curriculum stage & Cumulative total\\
\midrule
0 & 271,711 & 0.5 & Stages 2 \& 3 & 271,711\\
1 & 271,090 & 0.6 & Stage 3 only  & 542,801\\
2 & 269,880 & 0.7 & Stage 3 only  & 812,681\\
3 & 268,120 & 0.8 & Stage 3 only  & 1,080,801\\
4 & 265,460 & 0.9 & Stage 3 only  & 1,346,261\\
5 & 261,800 & 1.0 & Stage 3 only  & 1,608,061\\
\midrule
\textbf{Total} & \textbf{1,608,061} & 0.5 to 1.0 & & \\
\bottomrule
\end{tabular}
\end{table}

\begin{figure}[htbp]
\centering
\begin{tabular}{ccccc}
\includegraphics[width=0.14\textwidth]{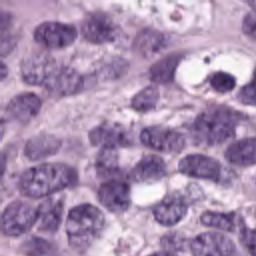} &
\includegraphics[width=0.14\textwidth]{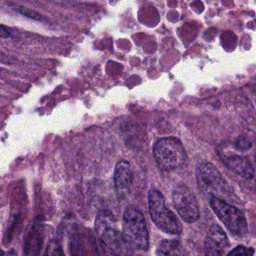} &
\includegraphics[width=0.14\textwidth]{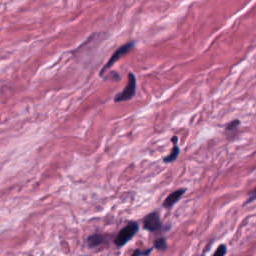} &
\includegraphics[width=0.14\textwidth]{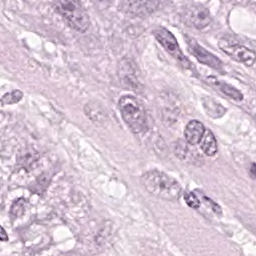} &
\includegraphics[width=0.14\textwidth]{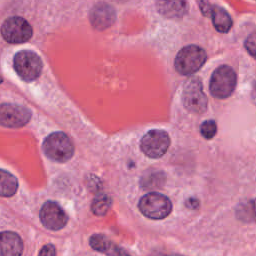} \\
Image 500 & Image 1000 & Image 1500 &
Image 2000 & Image 2500 
\end{tabular}
\caption{\textbf{Representative samples from PanNuke dataset - fold 3.}
PanNuke dataset was taken using 0.25\,$\mu$m/pixel, 
Each column shows the same 256$\times$256 pixel from 64$\times$64\,$\mu$m tissue area. PanNuke is a high-resolution nuclei dataset, in-contrast to TCGA-UT dataset which the tissue area is bigger, and taken using higher $\mu$m/pixel}
\label{fig:pannuke_samples}
\end{figure}


\subsection{\seg{}: Dual-Head Segmentation Architecture}
\label{sec:arch}

\subsubsection{Overview}

\begin{figure}[htbp]
\centering
\resizebox{\textwidth}{!}{%
\begin{tikzpicture}[
    font=\sffamily\footnotesize,
    node distance=1.5cm and 2cm,
    block/.style={
        rectangle, draw, rounded corners,
        minimum width=2.5cm, minimum height=1cm,
        align=center, fill=white, drop shadow},
    backbone/.style={
        rectangle, draw, fill=blue!10,
        minimum width=3cm, minimum height=8cm, align=center},
    layer/.style={
        rectangle, draw, fill=blue!30,
        minimum width=2.5cm, minimum height=0.6cm, align=center},
    neck/.style={
        rectangle, draw, fill=green!10,
        minimum width=2.5cm, minimum height=6cm,
        align=center, dashed},
    head/.style={
        rectangle, draw, fill=orange!20,
        minimum width=3cm, minimum height=1.5cm, align=center},
    output/.style={
        rectangle, draw, fill=red!10,
        minimum width=2cm, minimum height=2cm, align=center},
    connection/.style={-Stealth, thick, draw=gray!80},
    label/.style={text=gray!50!black, font=\scriptsize},
    rpbox/.style={
        rectangle, draw, rounded corners=4pt,
        minimum width=3.6cm, minimum height=0.85cm,
        align=center, font=\sffamily\scriptsize,
        inner sep=4pt, drop shadow},
    rpio/.style={
        trapezium, trapezium left angle=75, trapezium right angle=105,
        draw, minimum width=3.2cm, minimum height=0.85cm,
        align=center, font=\sffamily\scriptsize\bfseries,
        inner sep=4pt},
    rparr/.style={-{Stealth[length=3pt,width=3pt]}, thick, draw=gray!70},
    rpgrp/.style={
        rectangle, rounded corners=6pt, draw, dashed,
        inner sep=8pt},
]
 
 
\node[output, fill=gray!20] (input) {Input Image\\$3 \times 224 \times 224$};
 
\node[backbone, right=1.0cm of input] (vit) {};
\node[above, font=\sffamily\bfseries\footnotesize] at (vit.north)
    {UNI2-h Backbone};
\node[below, font=\sffamily\scriptsize, text width=2.5cm, align=center]
    at (vit.south) {ViT-Giant\\(Pretrained, fine-tuned)};
 
\node[layer, fill=blue!20] (b1)
    at ($(vit.south)!0.15!(vit.north)$) {Block 5 \\ (Scale 1/4)};
\node[layer, fill=blue!25] (b2)
    at ($(vit.south)!0.40!(vit.north)$) {Block 11 \\ (Scale 1/8)};
\node[layer, fill=blue!30] (b3)
    at ($(vit.south)!0.65!(vit.north)$) {Block 17 \\ (Scale 1/16)};
\node[layer, fill=blue!40] (b4)
    at ($(vit.south)!0.90!(vit.north)$) {Block 23 \\ (Scale 1/32)};
 
\draw[connection] (input) -- (vit);
 
\node[block, right=0.8cm of b1, minimum width=1.5cm, fill=yellow!20]
    (p1) {Proj $1\times1$\\256 ch};
\node[block, right=0.8cm of b2, minimum width=1.5cm, fill=yellow!20]
    (p2) {Proj $1\times1$\\512 ch};
\node[block, right=0.8cm of b3, minimum width=1.5cm, fill=yellow!20]
    (p3) {Proj $1\times1$\\1024 ch};
\node[block, right=0.8cm of b4, minimum width=1.5cm, fill=yellow!20]
    (p4) {Proj $1\times1$\\2048 ch};
 
\draw[connection] (b1) -- (p1);
\draw[connection] (b2) -- (p2);
\draw[connection] (b3) -- (p3);
\draw[connection] (b4) -- (p4);
 
\node[right=3.0cm of vit,
      minimum height=7cm, minimum width=2.5cm,
      dashed, draw=green!50!black, fill=green!5, rounded corners]
    (decoder) {};
\node[above, font=\sffamily\bfseries\footnotesize, text=green!50!black]
    at (decoder.north) {UperNet Decoder};
\node[font=\sffamily\scriptsize, align=center]
    at (decoder.center)
    {Pyramid Pooling\\Module (PPM)\\+\\FPN Fusion};
 
\draw[connection] (p1) -- (p1 -| decoder.west);
\draw[connection] (p2) -- (p2 -| decoder.west);
\draw[connection] (p3) -- (p3 -| decoder.west);
\draw[connection] (p4) -- (p4 -| decoder.west);
 
\node[head, right=1.0cm of decoder, yshift=2.0cm]
    (sem_head) {Semantic Head\\(6-class)};
\node[head, right=1.0cm of decoder, yshift=-2cm, fill=purple!20]
    (hv_head) {HV Head\\(Regression)};
 
\draw[connection, line width=1.5pt]
    (decoder.east) -- ++(0.5,0) |- (sem_head.west);
\draw[connection, line width=1.5pt]
    (decoder.east) -- ++(0.5,0) |- (hv_head.west);
 
\node[output, right=0.5cm of sem_head, fill=green!10]
    (sem_out) {Semantic Mask\\$6 \times H \times W$};
\node[output, right=0.5cm of hv_head, fill=purple!10]
    (hv_out) {HV Maps\\$2 \times H \times W$};
 
\draw[connection] (sem_head) -- (sem_out);
\draw[connection] (hv_head) -- (hv_out);
 
\node[below=0.1cm of sem_out, font=\sffamily\tiny, align=center]
    {Classes: Neo, Inf,\\Conn, Dead, Epi, BG};
\node[below=0.1cm of hv_out, font=\sffamily\tiny, align=center]
    {Channels:\\Horizontal $\Delta$, Vertical $\Delta$};
 
\node[below=1.1cm of decoder,
      font=\sffamily\itshape\tiny, text width=6cm, align=center]
    {$^*$Features at transformer blocks 5, 11, 17, 23\\
     correspond to hierarchical scales 1/4, 1/8, 1/16, 1/32.};
 
\node[above=1.5cm of vit, font=\sffamily\bfseries\small,
      text=blue!60!black]
    {(a)\ \ Model Architecture};
 
\begin{scope}[on background layer]
    \node[rpgrp, draw=blue!30, fill=blue!2,
          fit=(input)(vit)(p1)(p4)(decoder)(sem_head)(hv_head)
              (sem_out)(hv_out)] (leftpanel) {};
\end{scope}
 
\draw[gray!40, very thick, dashed]
    ($(leftpanel.north east)+(0.55,0.3)$)
    --
    ($(leftpanel.south east)+(0.55,-0.3)$);
 
 
\coordinate (rpstart) at
    ($(leftpanel.north east)+(3.4, 0)$);
 
\node[rpio, fill=gray!15, draw=gray!60]
    (rp_patch) at (rpstart)
    {Patch Image\\(H\&E, any size)};
 
\node[rpbox, fill=cyan!10, draw=cyan!60,
      below=0.55cm of rp_patch]
    (rp_mpp)
    {MPP Normalise\\$s = \mathrm{MPP}_\mathrm{input} \div 0.314$};
 
\node[rpbox, fill=cyan!10, draw=cyan!60,
      below=0.55cm of rp_mpp]
    (rp_tile)
    {Tile $224^2$\\ 50\% overlap (stride 112\,px)};
 
\node[rpbox, fill=blue!12, draw=blue!50,
      below=0.55cm of rp_tile,
      minimum width=3.6cm, minimum height=1.1cm]
    (rp_inf)
    {\textbf{UNI2-UperHoVer}\\Inference\\per tile};
 
\node[rpbox, fill=green!10, draw=green!55,
      below=0.55cm of rp_inf, xshift=-1.05cm,
      minimum width=1.55cm]
    (rp_stit_sem)
    {Stitch\\Semantic Map\\(fg-priority)};
 
\node[rpbox, fill=purple!10, draw=purple!50,
      below=0.55cm of rp_inf, xshift=1.05cm,
      minimum width=1.55cm]
    (rp_stit_hv)
    {Stitch\\HV Map\\(last-write)};
 
\node[rpbox, fill=orange!12, draw=orange!60,
      below=0.65cm of rp_stit_sem, xshift=1.05cm]
    (rp_ws)
    {Watershed\\Instance Map};
 
\node[rpbox, fill=yellow!20, draw=yellow!70!black,
      below=0.55cm of rp_ws]
    (rp_tme)
    {TME Feature\\Extraction (JSON)};
 
\node[rpbox, fill=teal!12, draw=teal!60,
      below=0.55cm of rp_tme]
    (rp_bio)
    {BioNeMo SFT\\GPT Model};
 
\node[rpio, fill=teal!20, draw=teal!60,
      below=0.55cm of rp_bio]
    (rp_out)
    {TME Narrative\\Output (text)};
 
\draw[rparr] (rp_patch) -- (rp_mpp);
\draw[rparr] (rp_mpp)   -- (rp_tile);
\draw[rparr] (rp_tile)  -- (rp_inf);
 
\draw[rparr] (rp_inf.south) -- ++(0,-0.28)
    -| (rp_stit_sem.north);
\draw[rparr] (rp_inf.south) -- ++(0,-0.28)
    -| (rp_stit_hv.north);
 
\draw[rparr] (rp_stit_sem.south) |- ++(0,-0.25)
    -| (rp_ws.north);
\draw[rparr] (rp_stit_hv.south)  |- ++(0,-0.25)
    -| (rp_ws.north);
 
\draw[rparr] (rp_ws)  -- (rp_tme);
\draw[rparr] (rp_tme) -- (rp_bio);
\draw[rparr] (rp_bio) -- (rp_out);
 
\node[right=0.25cm of rp_mpp, font=\sffamily\tiny,
      text=black!70, align=left, text width=2.2cm]
    {Scale to ${\approx}0.35\,\mu$m/px};
 
\node[right=0.25cm of rp_tile, font=\sffamily\tiny,
      text=black!70, align=left, text width=2.2cm]
    {Zero-pad to $\div14$};
 
\node[right=0.25cm of rp_ws, font=\sffamily\tiny,
      text=black!70, align=left, text width=2.2cm]
    {Energy + EDT seeds\\compactness 0.01};
 
\node[right=0.25cm of rp_tme, font=\sffamily\tiny,
      text=black!70, align=left, text width=2.2cm]
    {22 features:\\counts, fractions,\\spatial entropy,\\KD-tree distances};
 
\node[right=0.25cm of rp_bio, font=\sffamily\tiny,
      text=black!70, align=left, text width=2.2cm]
    {Structured JSON\\$\rightarrow$ clinical narrative};
 
\begin{scope}[on background layer]
    \node[rpgrp, draw=cyan!40, fill=cyan!3,
          fit=(rp_mpp)(rp_tile),
          label={[font=\sffamily\tiny\bfseries,
                  text=cyan!60!black, anchor=south west]
                  south west:Pre-processing}]
        (grp_pre) {};
 
    \node[rpgrp, draw=blue!30, fill=blue!2,
          fit=(rp_inf)(rp_stit_sem)(rp_stit_hv),
          label={[font=\sffamily\tiny\bfseries,
                  text=blue!60, anchor=south west]
                  south west:Model inference}]
        (grp_inf) {};
 
    \node[rpgrp, draw=orange!40, fill=orange!2,
          fit=(rp_ws)(rp_tme),
          label={[font=\sffamily\tiny\bfseries,
                  text=orange!70!black, anchor=south west]
                  south west:Post-processing}]
        (grp_post) {};
 
    \node[rpgrp, draw=teal!40, fill=teal!3,
          fit=(rp_bio)(rp_out),
          label={[font=\sffamily\tiny\bfseries,
                  text=teal!60!black, anchor=south west]
                  south west:LLM reporting}]
        (grp_llm) {};
\end{scope}
 
\node[above=0.5cm of rp_patch,
      font=\sffamily\bfseries\small, text=teal!60!black]
    {(b)\ \ Inference \& Reporting Pipeline};
 
\end{tikzpicture}
}
\caption{\textbf{Full \fw{} system architecture and inference pipeline.}
\textit{(a)} \seg{} dual-head segmentation model: a shared \bb{} ViT-Giant backbone
extracts a four-scale FPN from transformer blocks 5, 11, 17, and 23; two independent
UperNet decoder heads produce six-class semantic logits and two-channel HV maps.
\textit{(b)} End-to-end inference and reporting pipeline: patches are MPP-normalised,
tiled with 50\% overlap, processed by \seg{}, stitched, post-processed by watershed
instance segmentation, and fed into TME feature extraction and BioNeMo narrative
generation.}
\label{fig:arch}
\end{figure}


\Cref{fig:arch} presents the full \fw{} system as a two-panel diagram. The left panel~(a)
shows the \seg{} model architecture: a shared \bb{} ViT-Giant backbone feeds four-scale
FPN features into two independent UperNet decoder heads, one producing six-class semantic
logits and one producing two-channel HV gradient maps. The right panel~(b) illustrates the
end-to-end inference and reporting pipeline, from MPP normalisation and tiling through
watershed-based instance separation, TME feature extraction, and BioNeMo narrative
generation. The dual-head design is novel in that it enables simultaneous semantic
classification and nuclear instance separation from a single shared foundation model
encoder, removing the need for separate specialist models and enabling unified training
on the same patch without any additional instance-level ground truth beyond what can be
synthesised dynamically from semantic labels.

\subsubsection{\bb{} multi-scale feature extractor}

\bb{} is a ViT-Giant model pretrained on $>$200M histopathology tiles from 350,000 slides
across 40+ cancer types~\cite{chen2024uni2}. Configuration: patch size 14, embedding
dimension $d=1536$, depth $L=24$ transformer blocks, $H=24$ attention heads, SwiGLU-packed
MLP (ratio 5.33), 8 register tokens, no classification head.

Because ViT architectures produce a flat token sequence at a single stride, a four-scale
feature pyramid is constructed by extracting patch token sequences at blocks
$b\in\{5,11,17,23\}$ ($\approx\{25,50,75,100\}\%$ of network depth). These four taps
are chosen to correspond to roughly equal quartile depths of the 24-block network,
providing a balanced hierarchy of low-level, mid-level, high-level, and semantic
features respectively, which mirrors the design rationale of conventional CNN-based FPNs.
The feature extraction follows \Cref{eq:fpn}:
\begin{equation}
  \mathbf{F}^{(b)}=W^{(b)}_{1\times1}\,\text{Reshape}\!\left(
  \mathbf{x}_\text{patch}^{(b)}\right),\quad b\in\{5,11,17,23\}
  \label{eq:fpn}
\end{equation}
The $\text{Reshape}$ operation converts the flat patch token sequence
(shape $N_\text{tok}\times d$) back to a 2D spatial map (shape
$h_\text{feat}\times w_\text{feat}\times d$), and the $1\times1$ convolution
$W^{(b)}_{1\times1}$ projects from the backbone embedding dimension $d=1536$
to standard FPN channel counts $c_b\in\{256,512,1024,2048\}$ (\Cref{tab:fpn}).
Bilinear interpolation then aligns each feature map to its target FPN stride.
The resulting spatial dimensions for a $224\times224$-pixel input are shown in
\Cref{tab:fpn}; all four feature maps are subsequently consumed by both UperNet
decoder heads.

\begin{table}[htbp]
\centering
\caption{Multi-scale feature pyramid configuration extracted from \bb{}.}
\label{tab:fpn}
\small
\begin{tabular}{ccccc}
\toprule
Block & Depth (\%) & FPN stride & Channels & Spatial size (224-px input)\\
\midrule
5  & 25  & $s/4$  & 256  & $56\times56$\\
11 & 50  & $s/8$  & 512  & $28\times28$\\
17 & 75  & $s/16$ & 1024 & $14\times14$\\
23 & 100 & $s/32$ & 2048 & $7\times7$\\
\bottomrule
\end{tabular}
\end{table}

\FloatBarrier

\subsubsection{Dual UperNet decoder heads}

Both heads are instantiated as \texttt{UperNetForSemanticSegmentation} decoders
(Hugging Face Transformers~\cite{wolf2020transformers}), with the internal backbone
replaced by a null reference so only the \texttt{decode\_head} (PPM + FPN fusion +
segmentation convolution) is active. Parameters are not shared between the two
heads: the semantic and HV regression tasks have fundamentally different objectives
(discrete class probability estimation versus continuous gradient field regression),
and sharing parameters would force a single decoder to simultaneously optimise for
softmax-normalised six-class output and unbounded two-channel regression, degrading
performance on both.

Each head processes the four-scale FPN features in two stages. First, a Pyramid
Pooling Module (PPM) applies pooling at multiple spatial scales on the coarsest feature
map and concatenates the results, capturing global context that would otherwise be lost
at small spatial resolutions. Second, an FPN fusion step progressively combines feature
maps from coarser to finer scales via lateral connections, producing a single fused
representation that encodes both local detail and global context simultaneously.
The fused representation is then passed through a segmentation convolution and bilinearly
upsampled to the full input resolution for loss computation.

\paragraph{Semantic head.} \texttt{num\_labels=6}, \texttt{hidden\_size=768}.
Outputs $\hat{\mathbf{Y}}\in\mathbb{R}^{B\times6\times H\times W}$ (PanNuke ontology),
bilinearly upsampled to full resolution before loss computation.

\paragraph{HV regression head.} \texttt{num\_labels=2}, \texttt{hidden\_size=768}.
Outputs $\hat{\mathbf{M}}\in\mathbb{R}^{B\times2\times H\times W}$
(channel 0: horizontal gradient map; channel 1: vertical), bilinearly upsampled;
no final activation applied.

\subsubsection{Loss functions}

\paragraph{Semantic cross-entropy:}
Here $\Omega_v$ denotes the set of valid pixels (\ie{} all pixels with ground-truth label
$\neq255$; the ignore index 255 is used for boundary regions and unlabelled voids), and
$\hat{p}_{y_p}(p)$ is the predicted softmax probability assigned to the true class $y_p$
at pixel $p$. The semantic loss is defined in \Cref{eq:sem}:
\begin{equation}
  \loss{sem} = -\frac{1}{|\Omega_v|}\sum_{p\in\Omega_v}\log\hat{p}_{y_p}(p),
  \qquad \text{ignore\_index}=255
  \label{eq:sem}
\end{equation}

\paragraph{HoVer regression loss:}
Operating exclusively on foreground pixels ($y_p>0$) with foreground mask
$\mathbf{M}_\text{fg}$, the HV regression loss combines a pixel-wise MSE term
(\Cref{eq:mse}) with a gradient-matching term (\Cref{eq:msge}):
\begin{align}
  \loss{MSE}  &= \tfrac{1}{|\Omega_\text{fg}|}
    \lVert(\hat{\mathbf{M}}-\mathbf{T})\odot\mathbf{M}_\text{fg}\rVert_F^2
    \label{eq:mse}\\
  \loss{MSGE} &= \tfrac{1}{|\Omega_\text{fg}|}\!\Bigl[
    \lVert(\nabla_x\hat{M}_0-\nabla_xT_0)\odot M_\text{fg}\rVert_F^2
    +\lVert(\nabla_y\hat{M}_1-\nabla_yT_1)\odot M_\text{fg}\rVert_F^2\Bigr]
    \label{eq:msge}\\
  \loss{hv}   &= \loss{MSE}+2\cdot\loss{MSGE}
    \label{eq:hv}
\end{align}
Here $\nabla_x$ and $\nabla_y$ are finite-difference spatial gradient operators applied
along the horizontal and vertical axes respectively (implemented as Sobel-style
convolution with \texttt{torch.nn.functional.conv2d}). The gradient term
$\loss{MSGE}$ penalises errors in the \emph{slope} of the predicted HV maps near nuclear
boundaries, where the gradient magnitude is largest and correct slope direction is most
critical for watershed energy surface construction. The weight of 2 on $\loss{MSGE}$
reflects the importance of boundary sharpness over absolute HV value accuracy.

\paragraph{Combined objective:}
\begin{equation}
  \loss{total}=\loss{sem}+\lambda\,\loss{hv},\qquad\lambda=1.0
  \label{eq:total}
\end{equation}
The choice $\lambda=1.0$ balances semantic and HV objectives without requiring separate
hyperparameter tuning: both loss components operate on normalised quantities (cross-entropy
for semantics; MSE in the $[-1,1]^2$ HV space), so their magnitudes are naturally
comparable. This choice is adopted from the original HoVer-Net formulation~\cite{graham2019hover}
and held fixed across all three curriculum stages; a formal ablation of $\lambda$ is
planned for the companion paper.

\subsubsection{Dynamic HV target synthesis}

Instance-level polygon annotations (one mask polygon per nucleus) are expensive to
produce at scale: PanNuke required trained pathologists to delineate 189,744 individual
nuclei across 7,901 images, yet this effort covers only a single institution's staining
protocol at a single resolution. Semantic-only annotations (a class label per pixel,
without separating touching nuclei of the same class) are substantially cheaper and are
the natural output of pseudo-labelling. Dynamic HV target synthesis bridges this gap by
deriving approximate per-nucleus HV gradient targets directly from connected-component
analysis of semantic masks, without any instance-level ground truth:
\begin{enumerate}[leftmargin=*, noitemsep, topsep=2pt]
  \item Binarise the semantic map ($y>0$); apply connected-component labelling
        (\texttt{scipy.ndimage.measurements.label}).
  \item Compute centroid $(\hat{c}_y,\hat{c}_x)$ of each component via
        \texttt{center\_of\_mass}.
  \item For pixel $(y,x)$ in component $i$:
        $T_x=\mathrm{clip}\!\left(\tfrac{x-\hat{c}_x}{10},-1,1\right)$,
        $T_y=\mathrm{clip}\!\left(\tfrac{y-\hat{c}_y}{10},-1,1\right)$
\end{enumerate}
The clip function constrains the centroid-relative distance to the range $[-1,1]$
using a scaling factor of 10 pixels, so that the HV target saturates at $\pm1$ for
pixels more than 10 pixels from the centroid; this prevents the gradient surface from
growing unboundedly for large nuclei and makes the target range consistent with the
$\tanh$-like energy surfaces used in the watershed post-processing step.
This procedure runs on CPU inside \texttt{torch.no\_grad()} and adds negligible latency.
Critically, because the synthesis requires only a semantic mask as input, pseudo-label
training in Stages~2 and~3 can generate both semantic and HV targets for the full
1.6M-patch TCGA-UT corpus without any instance-level human annotations, enabling
the progressive curriculum to scale to the full TCGA diversity.

\subsection{Three-Stage Progressive Pseudo-Label Curriculum}
\label{sec:curriculum}

\begin{figure}[t]
\centering
\includegraphics[width=0.8\textwidth]{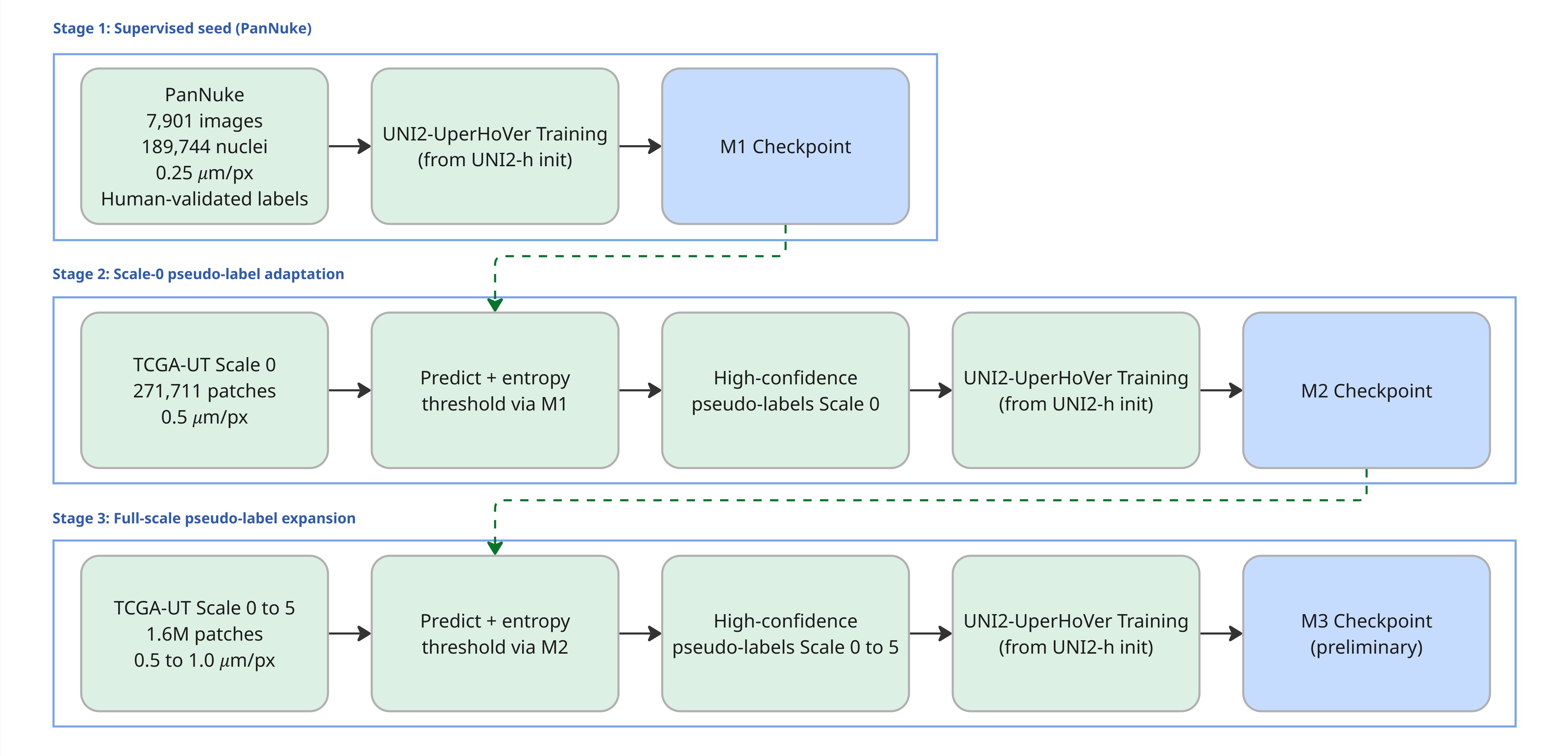}
\caption{\textbf{Three-stage progressive pseudo-label training curriculum.}
Each model is trained independently from a fresh initialisation (\bb{} pretrained
backbone, randomly initialised decoder heads); the only connection between stages
is the pseudo-labels generated by the previous model.
\textbf{Stage 1:} $\mathcal{M}_1$ is trained on human-annotated PanNuke (0.25\,$\mu$m/pixel).
\textbf{Stage 2:} $\mathcal{M}_1$ generates entropy-filtered pseudo-labels for
TCGA-UT scale~0 (271,711 patches; 0.5\,$\mu$m/pixel); $\mathcal{M}_2$ is trained
from fresh initialisation on these pseudo-labels.
\textbf{Stage 3:} The TCGA-adapted $\mathcal{M}_2$ generates entropy-filtered
pseudo-labels for all six TCGA-UT scales (1.6M~patches; 0.5 to 1.0\,$\mu$m/pixel);
$\mathcal{M}_3$ is trained from fresh initialisation on these pseudo-labels and
is the preliminary Stage~3 checkpoint; scale-aware retraining is required before
$\mathcal{M}_3$ can be considered validated (see \Cref{sec:discussion}).
Dashed arrows = pseudo-label generation (inference only); solid = training.}
\label{fig:curriculum}
\end{figure}

\Cref{fig:curriculum} summarises the three-stage training curriculum; subsequent
subsections describe each stage in detail.



\subsubsection{Training Configuration: Pre-processing and augmentation}
\label{sec:training}

Images are padded to the nearest multiple of 14 (right/bottom edges, zero-fill),
resized to $224\times224$ (bilinear for images; nearest-neighbour for masks), and
normalised with ImageNet statistics
($\mu=(0.485, 0.456, 0.406)$, $\sigma=(0.229, 0.224, 0.225)$).
Training augmentations (each applied with probability 0.5):
colour jitter (brightness, contrast, saturation $\in[0.8,1.2]$; hue $\in[-0.05,0.05]$);
HLS-space multiplicative perturbation $\in[0.9,1.1]$; horizontal flip; vertical flip.
Spatial augmentations are applied identically to image and mask.

\subsubsection{Training Configuration: Hyperparameters}
\Cref{tab:hparams} lists hyperparameters shared across all three stages.
Stage~1 and Stage~2 ran for 250 epochs. Stage~3 was trained for 100 epochs;
because each epoch processes all 1.6M~patches, Stage~3 accumulates
substantially more total gradient steps than the earlier stages despite the
lower epoch count (335,100 steps for Stage~3 vs.~24,750 for Stage~1 and 212,500 for
Stage~2, computed over each stage's 80\mbox{\%} training partition at its respective
effective batch size of 64, 256, and 384 sequences).

\begin{table}[htbp]
\centering
\caption{Training hyperparameters for all \seg{} curriculum stages. Parameters that
vary by stage are listed per stage. Reported step counts (\Cref{sec:training_dynamics})
reflect the 80\mbox{\%} training partition of each stage's corpus (20\mbox{\%} held out for
validation); effective batch size is per-device batch size \mbox{$\times$} GPU count
\mbox{$\times$} gradient accumulation steps as listed below.}
\label{tab:hparams}
\small
\begin{tabular}{ll}
\toprule
Hyperparameter & Value\\
\midrule
Optimiser                    & AdamW ($\beta_1=0.9$, $\beta_2=0.999$, weight\_decay=$1\times10^{-2}$)\\
Learning rate                & $5\times10^{-5}$\\
LR scheduler                 & Linear decay: $\text{LR}(t) = 5\times10^{-5}\cdot(T-t)/T$ (no warmup)\\
Per-device batch size        & 8\\
Gradient accumulation steps  & 1 (Stage~1); 4 (Stage~2); 8 (Stage~3)\\
Training epochs per stage    & 250 (Stage~1); 250 (Stage~2); 100 (Stage~3, 1.6M~patches/epoch)\\
Mixed precision              & bfloat16\\
Compilation                  & \texttt{torch.compile} (inductor backend)\\
Multi-GPU strategy           & DDP (\texttt{ddp\_timeout}=10,800\,s)\\
Number of GPUs               & 8 (Stages~1 and~2); 6 (Stage~3)\\
Evaluation frequency         & Every 500 steps\\
Checkpoint selection metric  & Validation mean IoU ($\uparrow$)\\
Checkpoints retained         & Best 2 per stage\\
Entropy threshold $\tau$     & 70th percentile of per-patch $\bar{H}$\\
HV loss weight $\lambda$     & 1.0\\
\bottomrule
\end{tabular}
\end{table}

\FloatBarrier

\subsection{TME Feature Extraction Pipeline}
\label{sec:tme}

\Cref{fig:tme_flowchart} shows the end-to-end TME feature extraction pipeline. There are two phases, described as follows.

\subsubsection{Phase~1 (\texttt{analyze\_tme\_from\_maps}, blue region)}

Validated masks are parsed via \texttt{skimage.measure.regionprops} to extract
per-nucleus centroid coordinates and pixel-area values, which are mapped to their
semantic class identities using \texttt{CLASS\_MAP} (linking instance IDs to
classes 0--5: Background, Neoplastic, Inflammatory, Connective, Dead, Epithelial).
Composition metrics are subsequently computed for each class: absolute nucleus count
$N_c$, population fraction $r_c = N_c / N_\text{total}$, and mean nuclear area
$\bar{a}_c$, along with two bounded clinical ratios, namely the Tumour-Stroma fraction
$f_\text{TS} = N_\text{neo} / (N_\text{neo} + N_\text{conn})$ and the Immune
Infiltration fraction $f_\text{II} = N_\text{inf} / (N_\text{neo} + N_\text{inf})$. Tumour-stroma fraction is used here to distinguish our automated, segmentation-derived metric from the manually-scored tumour-stroma ratio (TSR) of conventional histopathology. For Advanced Spatial Analysis step, two \texttt{scipy.spatial.cKDTree} structures are constructed over cell centroids.

The first tree, built over neoplastic centroids, is queried once per inflammatory cell to yield three immune-to-tumour spatial metrics: \texttt{min\_dist\_Inflammatory\_to\_Neoplastic} (closest tumour cell distance),
\texttt{mean\_dist\_Inflammatory\_to\_Neoplastic} (mean over all inflammatory cells),
and \texttt{count\_engaged\_immune} (number of inflammatory cells within a 50\,px
proximity threshold, quantifying direct immune-to-tumour contact).

The second tree, built over all cell centroids, retrieves the $k=6$ nearest neighbours
per cell; the Shannon entropy $H_i = -\sum_c p_{i,c} \log_2 p_{i,c}$ of each cell's
neighbour class distribution is computed and averaged across the patch to yield
\texttt{mean\_spatial\_entropy} that is a scalar index of cellular mixing, where high
values indicate heterogeneous immune-to-tumour co-localisation and low values indicate
class-segregated spatial compartmentalisation.

The 22 numeric features are collected into a \texttt{tme\_features} Python dictionary,
serving as the intermediate handoff between phases. The features are divided into four distinct categories: compositional features, clinical TME ratios, spatial interaction metrics, and spatial entropy; defined as follows:
\begin{enumerate}

\item{\textbf{Compositional features:}}
For each class $c\in\{1,\ldots,5\}$: absolute count $N_c$, population fraction
$r_c=N_c/N_\text{total}$, and mean nuclear area $\bar{a}_c$ (\texttt{regionprops}).

\item{\textbf{Clinical TME ratios:}}
The Tumour-Stroma fraction $f_\text{TS}$ and Immune Infiltration fraction $f_\text{II}$
are defined in \Cref{eq:ratios}:
\begin{equation}
  f_\text{TS}=\frac{N_\text{neo}}{N_\text{neo}+N_\text{conn}},\qquad
  f_\text{II}=\frac{N_\text{inf}}{N_\text{neo}+N_\text{inf}}
  \label{eq:ratios}
\end{equation}

\item{\textbf{Spatial interaction metrics:}}
A \texttt{scipy.spatial.cKDTree} over neoplastic centroids is queried with each
inflammatory centroid to return the single nearest tumour-cell distance.
Derived features: \texttt{mean\_dist\_Inflammatory\_to\_Neoplastic},
\texttt{min\_dist\_Inflammatory\_to\_Neoplastic}, and
\texttt{count\_engaged\_immune} (threshold 50\,px).

\item{\textbf{Spatial entropy (a cellular mixing index):}}
The per-cell Shannon entropy $H_i$ and its patch-level average $\bar{H}$ are defined in
\Cref{eq:spatial_entropy}:
\begin{equation}
  H_i=-\sum_c p_{i,c}\log_2 p_{i,c},\qquad
  \bar{H}=\frac{1}{N}\sum_{i=1}^{N}H_i
  \label{eq:spatial_entropy}
\end{equation}
using the six nearest neighbours of each cell as the local neighbourhood.
High $\bar{H}$ indicates heterogeneous cellular mixing (possible intra-tumoral
infiltration); low $\bar{H}$ indicates class-segregated spatial organisation.
\end{enumerate}

\subsubsection{Phase~2 (\texttt{translate\_tme\_to\_bio\_tokens}, yellow region)}

Biological logic thresholds are applied across six phenotype axes:
(i)~\textbf{Composition}: tumour purity level and dominant non-tumour cell type
(Immune-dominant vs.\ Stroma-dominant when $r_\text{neo} < 0.30$);
(ii)~\textbf{Immune status}: Immune Hot ($r_\text{inf} > 0.30$), Immune Cold
($r_\text{inf} < 0.10$), or Patchy infiltrate;
(iii)~\textbf{Stromal status}: Desmoplastic ($r_\text{conn} > 0.50$) or Stroma-Free
($r_\text{conn} < 0.05$);
(iv)~\textbf{Spatial interaction}: Direct Contact
($\bar{d}_{\text{inf} \to \text{neo}} < 20$\,px), Excluded ($\bar{d} > 60$\,px),
or Proximal;
(v)~\textbf{Architecture}: Highly Mixed ($\bar{H} > 0.70$) or Segregated
($\bar{H} < 0.40$);
(vi)~\textbf{Morphology}: Large/Pleomorphic ($\bar{a}_\text{neo} > 300\,\text{px}^2$)
or Small/Monotonous ($\bar{a}_\text{neo} < 150\,\text{px}^2$).

Each threshold rule emits one or more biological token strings (\textit{e.g.}
``Lymphocyte-rich'', ``Immune exclusion'', ``Nuclear enlargement'') forming a flat,
ordered token list.

The final output is a structured JSON object with three fields:
\texttt{raw\_data} (all numeric TME features), \texttt{interpretation} (six categorical
phenotype labels), and \texttt{generated\_tokens} (the flat biological token list),
providing a machine-readable and human-interpretable per-patch TME representation
suitable for downstream BioNeMo narrative generation, retrieval-augmented pathology
analysis, or patient survival modelling.


\begin{figure}[!htb]
\centering
\resizebox{0.45\textwidth}{!}{

\begin{tikzpicture}[
    node distance=1cm and 2cm,
    box/.style={rectangle, draw, rounded corners, fill=brown!5, text width=7.5cm, align=center, minimum height=1cm, thick},
    io/.style={trapezium, trapezium left angle=75, trapezium right angle=105, draw, fill=green!10, text width=6.5cm, align=center, minimum height=1cm, thick},
    logic_box/.style={rectangle, draw, rounded corners, fill=orange!5, text width=8.5cm, align=left, minimum height=1cm, thick, inner sep=10pt},
    arrow/.style={thick, -{Stealth[length=2.5mm, width=2.5mm]}},
    group_box/.style={rectangle, draw, dashed, thick, inner sep=15pt}
]

\node (input) [io] {\textbf{Input}\\ \texttt{inst\_mask} (2D IDs), \texttt{cls\_mask} (0-5)};


\node (parse) [box, below=of input] {\textbf{Parse Instance Data}\\Extract centroids and pixel areas via \texttt{regionprops}; map to \texttt{CLASS\_MAP}.};

\node (comp) [box, below=of parse] {\textbf{Compute Composition}\\Calculate Counts, Mean Areas, and Bounded Fractions (e.g., Tumor/Stroma).};

\node (spatial) [box, below=of comp] {\textbf{Advanced Spatial Analysis}\\Use \texttt{cKDTree} to compute:\\1. Min/Mean distances (Immune $\rightarrow$ Tumor)\\2. Local Shannon Entropy (Cell mixing)};

\node (tme_dict) [io, below=of spatial] {\textbf{Intermediate Output}\\ \texttt{tme\_features} (Dictionary)};


\node (logic) [logic_box, below=of tme_dict] {
    \textbf{Apply Biological Logic Thresholds:}\\[1ex]
    $\bullet$ \textbf{Composition:} Tumor Purity \& Dominance\\
    $\bullet$ \textbf{Immune:} Hot, Cold, or Patchy infiltrate\\
    $\bullet$ \textbf{Stroma:} Desmoplastic vs. Stroma-Free\\
    $\bullet$ \textbf{Interaction:} Direct Contact, Proximal, or Excluded\\
    $\bullet$ \textbf{Architecture:} Highly Mixed vs. Segregated\\
    $\bullet$ \textbf{Morphology:} Pleomorphic vs. Monotonous
};

\node (output) [io, below=of logic] {\textbf{Final Output}\\Bio Tokens JSON\\(\texttt{raw\_data} + \texttt{interpretation})};

\draw [arrow] (input) -- (parse);
\draw [arrow] (parse) -- (comp);
\draw [arrow] (comp) -- (spatial);
\draw [arrow] (spatial) -- (tme_dict);
\draw [arrow] (tme_dict) -- (logic);
\draw [arrow] (logic) -- (output);

\begin{scope}[on background layer]
    \node[group_box, fill=blue!5, fit=(parse)(comp)(spatial), 
          label={[anchor=south east, font=\bfseries, text=black!70]south east:\texttt{analyze\_tme\_from\_maps}}] (phase1) {};
          
    \node[group_box, fill=yellow!10, fit=(logic), 
          label={[anchor=south east, font=\bfseries, text=black!70]south east:\texttt{translate\_tme\_to\_bio\_tokens}}] (phase2) {};
\end{scope}

\end{tikzpicture}

}
\caption{\textbf{End-to-end pipeline for per-patch TME feature extraction and
biological token translation, implemented across two functional modules.}
The pipeline accepts two segmentation outputs: a 2D instance mask
(\texttt{inst\_mask}, integer region IDs) and a six-class semantic mask
(\texttt{cls\_mask}, values 0 to 5) and produces a structured JSON object
containing all quantitative TME descriptors and their categorical interpretations.
}
\label{fig:tme_flowchart}
\end{figure}

\FloatBarrier

\subsection{Biological Token Translation and BioNeMo Narrative Generation}
\label{sec:bionemo}

With \seg{} providing six-class semantic segmentation maps and watershed-derived nuclear
instance masks (Section~\ref{sec:arch}), and a three-stage curriculum ensuring that these
outputs are available across the full resolution range of the TCGA-UT corpus
(Section~\ref{sec:curriculum}), the next step is to translate the raw segmentation output
into clinically actionable information. Section~\ref{sec:tme} described how structured
per-patch TME features are computed from these segmentation outputs. This section
describes how those quantitative features are first encoded as categorical biological
tokens and then used to drive a language model that generates clinically plausible
patch descriptions (pending formal quantitative evaluation; see Section~\ref{sec:discussion}).

\subsubsection{Rule-based biological token translation}
Raw numeric TME features are mapped to categorical phenotype labels and a flat
biological token list (\Cref{tab:tokens}) via the \texttt{translate\_tme\_to\_bio\_tokens}
function. This translation (i)~provides an immediately interpretable categorical patch
summary, and (ii)~anchors the BioNeMo prompt to quantitatively grounded descriptors,
substantially reducing LLM hallucination risk.

Rule-based thresholds are used in preference to learned classifiers for three reasons:
interpretability (each threshold follows established tumour microenvironment spatial
biology conventions where available, making the token assignment auditable),
reproducibility (the same feature values always produce the same tokens without
stochastic elements), and the absence of any additional training data requirement
(no labelled feature-to-token examples are needed). Phenotype thresholds follow
established conventions from tumour microenvironment spatial biology: immune status
follows the hot/cold tumour framework~\cite{galon2006type,galon2019approaches},
architecture follows the cold/mixed/compartmentalized categorisation of
Keren~et~al.~\cite{keren2018structured}, stromal status follows percentage-based
tumour stroma quantification~\cite{bengtsson2026digital}, spatial interaction
follows established distance-based infiltration/exclusion
analysis~\cite{saltz2018spatial}, and morphology follows area-based nuclear
grading~\cite{kronqvist1998morphometric}. Composition thresholds and the specific
numeric cut-off values throughout were calibrated to this study's pixel-level
resolution and six-class ontology. It is not directly transferred from prior work,
which typically uses different units, magnifications, or scoring systems.
The six phenotype axes (Composition, Immune, Stromal, Spatial, Architecture, Morphology)
cover the primary TME descriptors used in clinical pathology reporting and correspond
to the dimensions routinely assessed in manual TME scoring. The token list is deliberately
flat and ordered to remain compatible with any text encoder downstream, including
bag-of-words retrievers, transformer sequence encoders, and the LLaMA-3.2-1B model fine-tuned here via BioNeMo.

A key design advantage of this approach is that the biological tokens and the resulting
narrative description are derived from aggregate patch-level statistics, specifically
total cell counts, population fractions, mean nuclear areas, and spatial distances,
rather than from the precise delineation of individual cell boundaries. This property
makes TME characterisation robust to imperfect segmentation. Even in challenging
cases where staining or scanner conditions hinder accurate cell segmentation, the
overall composition ratios and spatial interaction metrics computed from the ensemble
of detected cells remain statistically representative of the true TME phenotype,
as demonstrated across the inter-scanner variability of the IGNITE tiles
(\Cref{sec:case_study}). The generated token set can therefore retain clinical
accuracy at the patch level even when individual cell boundaries are imperfectly
resolved.

\begin{table}[htbp]
\centering
\caption{%
  Biological token translation rules mapping raw numeric TME features to
  categorical interpretation labels and clinical vocabulary tokens.
  Rules are applied in the order shown; Spatial Interaction engagement tokens
  ($\dagger$) are appended to the token list produced by the primary
  distance condition.
  Variable definitions:
  $r_\text{neo}$~=~\texttt{ratio\_Neoplastic},
  $r_\text{inf}$~=~\texttt{ratio\_Inflammatory},
  $r_\text{conn}$~=~\texttt{ratio\_Connective},
  $\bar{d}$~=~\texttt{mean\_dist\_Inflammatory\_to\_Neoplastic} (px),
  $e$~=~\texttt{count\_engaged\_immune}~/~\texttt{count\_Inflammatory},
  $n_\text{inf}$~=~\texttt{count\_Inflammatory},
  $\bar{H}$~=~\texttt{mean\_spatial\_entropy},
  $\bar{a}_\text{neo}$~=~\texttt{mean\_area\_Neoplastic} (px$^2$). Phenotype axis conventions
follow established TME spatial biology literature (see \Cref{sec:bionemo}
for citations); absolute numeric cut-offs were calibrated to this study's
pixel-level resolution.
}
\label{tab:tokens}
\small
\renewcommand{\arraystretch}{1.25}
\begin{tabularx}{\textwidth}{%
  >{\raggedright\arraybackslash}p{2.6cm}   
  >{\raggedright\arraybackslash}p{5.0cm}   
  >{\raggedright\arraybackslash}p{3.2cm}   
  >{\raggedright\arraybackslash}X           
}
\toprule
\textbf{Phenotype axis} &
\textbf{Condition} &
\textbf{Interpretation} &
\textbf{Label / token(s)} \\
\midrule

\multirow{4}{2.6cm}{Composition \newline (Tumour purity)}
  & $r_\text{neo} > 0.60$
  & High Tumour Purity
  & ``Solid tumor nest''; ``Hypercellular'' \\

  & $r_\text{neo} < 0.30$ and $r_\text{inf} > r_\text{conn}$
  & Low Tumour Purity
  & ``Immune-dominant'' \\

  & $r_\text{neo} < 0.30$ and $r_\text{inf} \leq r_\text{conn}$
  & Low Tumour Purity
  & ``Stroma-dominant'' \\

  & $0.30 \leq r_\text{neo} \leq 0.60$
  & Moderate Cellularity
  & (no tokens) \\

\midrule

\multirow{3}{2.6cm}{Immune status \newline (Hot / Cold)}
  & $r_\text{inf} > 0.30$
  & Immune Hot
  & ``Lymphocyte-rich''; ``Heavy infiltrate'' \\

  & $r_\text{inf} < 0.10$
  & Immune Cold
  & ``Immune desert''; ``Pauci-immune'' \\

  & $0.10 \leq r_\text{inf} \leq 0.30$
  & Moderate Infiltration
  & ``Patchy infiltrate'' \\

\midrule

\multirow{3}{2.6cm}{Stromal status \newline (Desmoplasia)}
  & $r_\text{conn} < 0.05$
  & Stroma-Free
  & ``Non-desmoplastic''; ``Lack of fibrosis'' \\

  & $r_\text{conn} > 0.50$
  & Desmoplastic
  & ``Sclerotic stroma''; ``Fibrotic barrier'' \\

  & $0.05 \leq r_\text{conn} \leq 0.50$
  & Normal Stroma
  & (no tokens) \\

\midrule

\multirow{6}{2.6cm}{Spatial interaction \newline (Proximity \& engagement)}
  & $\bar{d} = -1$ (no cells of either type present)
  & No Interaction
  & ``Lacks interacting cell populations'' \\

  & $\bar{d} < 20\,\text{px}$
  & Direct Contact
  & ``Intra-tumoral infiltration''; ``Juxtaposed cells''; ``Cytotoxic potential'' \\

  & $\bar{d} > 60\,\text{px}$
  & Excluded
  & ``Immune exclusion''; ``Restricted to stroma'' \\

  & $20\,\text{px} \leq \bar{d} \leq 60\,\text{px}$
  & Proximal
  & ``Peritumoral'' \\

  & $e > 0.80$ \quad ($\dagger$engagement nuance)
  & (appended to primary)
  & ``Maximal immune engagement'' \\

  & $e < 0.20$ and $n_\text{inf} > 10$ \quad ($\dagger$)
  & (appended to primary)
  & ``Bystander immune cells'' \\

\midrule

\multirow{3}{2.6cm}{Architecture \newline (Spatial entropy)}
  & $\bar{H} > 0.70$
  & High Mixing
  & ``Diffuse integration''; ``Heterogeneous distribution'' \\

  & $\bar{H} < 0.40$
  & Segregated
  & ``Homogeneous'' \\

  & $0.40 \leq \bar{H} \leq 0.70$
  & Intermediate / Moderately Mixed
  & ``Moderately Mixed'' \\

\midrule

\multirow{4}{2.6cm}{Morphology \newline (Nuclear area)}
  & $\bar{a}_\text{neo} > 300\,\text{px}^2$
  & Large / Pleomorphic
  & ``Nuclear enlargement''; ``Anisocytosis'' \\

  & $0 < \bar{a}_\text{neo} < 150\,\text{px}^2$
  & Small / Monotonous
  & ``Small cell morphology'' \\

  & $150 \leq \bar{a}_\text{neo} \leq 300\,\text{px}^2$
  & (no explicit label)
  & (no tokens) \\

  & $\bar{a}_\text{neo} = 0$ (no neoplastic cells)
  & N/A
  & (no tokens) \\

\bottomrule
\end{tabularx}
\end{table}

\FloatBarrier

\subsubsection{LLaMA-3.2-1B supervised fine-tuning via BioNeMo}

\begin{figure}[htbp]
\centering
\includegraphics[width=\linewidth]{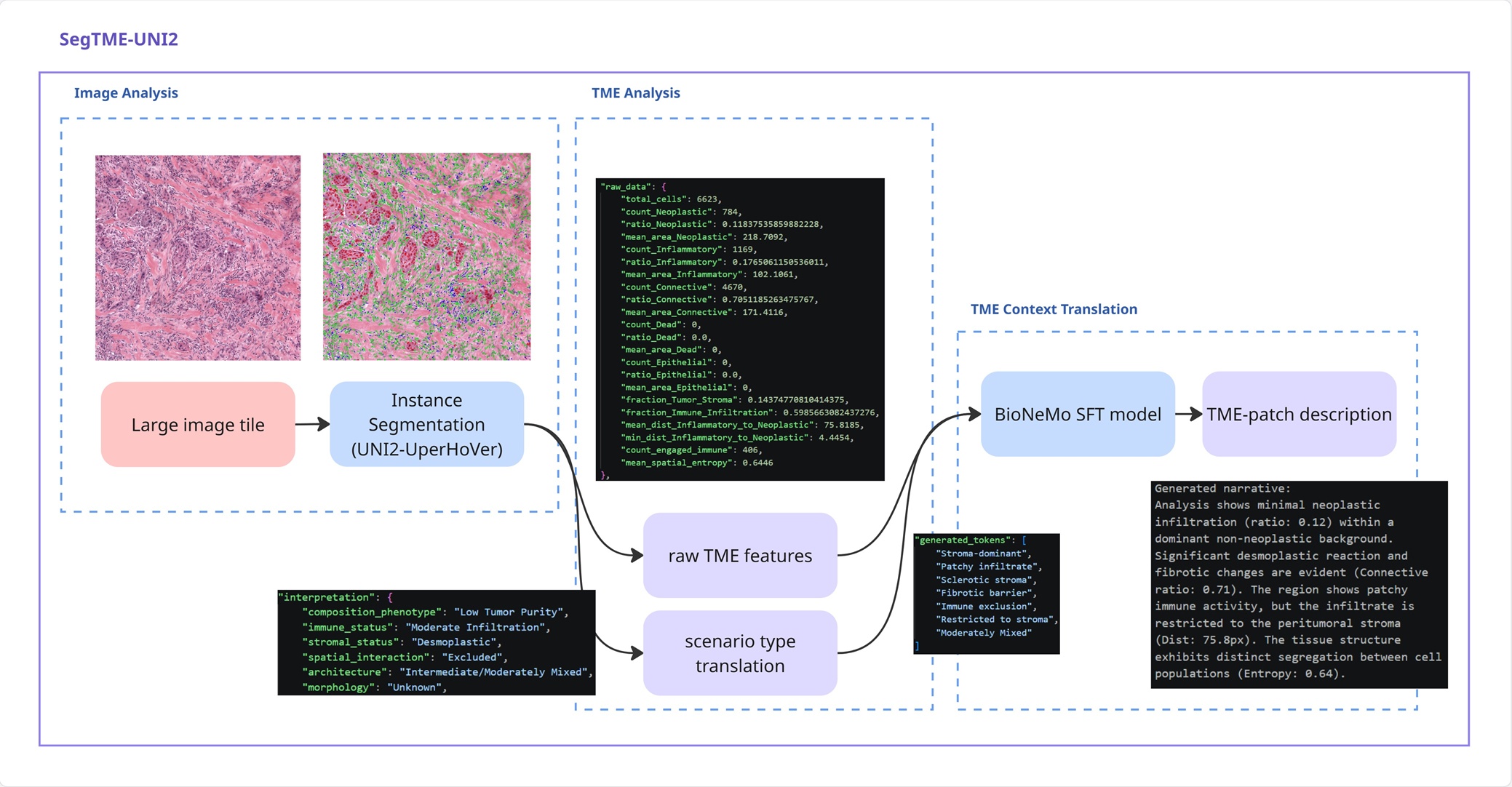}
\caption{\textbf{End-to-end \fw{} inference and narrative generation pipeline.}
From a raw H\&E patch, \seg{} produces a numeric TME feature set (raw\_data, green)
and categorical interpretation labels (orange). Both are passed as structured JSON
to the BioNeMo SFT model, which generates a clinically grounded TME patch description.
The representative example shown (Neoplastic ratio: 0.29; Immune Hot; immune cells
spatially excluded, mean distance: 136.1\,px; entropy: 0.27) illustrates the complete
pipeline from pixel prediction to natural language pathology reporting.}
\label{fig:bionemo_pipeline}
\end{figure}

The structured text prompt, serialised from \texttt{raw\_data} (numeric features),
\texttt{interpretation} (categorical labels from \Cref{tab:tokens}), and
\texttt{generated\_tokens} (flat token list), is used to fine-tune
\texttt{meta-llama/Llama-3.2-1B} (\texttt{NVLlamaForCausalLM}) via NVIDIA BioNeMo's
TransformerEngine-native SFT recipe. The fine-tuned model generates clinically
interpretable per-patch TME narratives, as shown in \Cref{fig:bionemo_pipeline}.

Supervised Fine-Tuning (SFT) in this context means that LLaMA-3.2-1B, pre-trained
by Meta on general-domain text, is further fine-tuned on paired examples consisting
of a structured text prompt (the feature set computed by the TME pipeline) and
a target narrative output (a clinical-style description of the TME phenotype).
Pretrained weights are loaded and converted from HuggingFace format to the
TransformerEngine fused-kernel tensor layout
before fine-tuning commences, ensuring the full benefit of pre-trained linguistic
knowledge is retained. During inference, the fine-tuned model takes a new structured
prompt as input and autoregressively generates the corresponding narrative,
conditioned on the numeric values it receives.

\paragraph{Training corpus and narrative provenance.}
Target narratives for the 1.6M training pairs were generated by a
template-based expansion of the rule-based labels and tokens (\Cref{sec:tme}),
using \texttt{random.choice()} over a pool of three surface-form phrasings per
phenotype dimension (\textit{e.g.}\ three ways to describe \textit{Immune Hot},
three ways to describe \textit{Excluded}, etc.) to provide lexical diversity across
the 1.6M records while keeping every clinical claim fully determined by the upstream
rule-based thresholds. No per-patch pathologist annotation was used; the SFT corpus
is entirely synthetic. The 50,000 training steps at effective batch size 32
correspond to approximately one pass over the full training corpus
($50{,}000 \times 32 = 1{,}600{,}000 \approx 1{,}608{,}061$ examples).

\paragraph{Prompt format.}
The four information levels are serialised into a single flat text prompt:
\begin{quote}\small
\texttt{METADATA | CANCER:}~\textit{type}~\texttt{| RES:}~\textit{res}~\texttt{um/px |}\\
\texttt{COORDS:}~\textit{coords}~\texttt{| MARGIN:}~\textit{margin}~\texttt{|
OUTCOME:}~\textit{outcome}~\texttt{||}\\
\texttt{RAW\_FEATURES: [}$k$\texttt{:}$v$\texttt{ pairs of all numeric TME fields]}\\
\texttt{SCENARIO: [}$k$\texttt{:}$v$\texttt{ pairs of six categorical labels]}\\
\texttt{KEYWORDS: [}$t_1, t_2, \ldots$\texttt{]}
\end{quote}
The \texttt{METADATA} block carries patch provenance (cancer type and physical
resolution from the folder-ID map), together with three reserved fields,
\texttt{COORDS}, \texttt{MARGIN}, and \texttt{OUTCOME}, populated as \texttt{NONE} in
this corpus since TCGA-UT patch-level clinical metadata is not available. These
fields are reserved for a planned follow-on training pass over WSI-derived data, where
they would carry a tile's slide coordinates, its margin status, and patient outcome
respectively. The segmentation backbone already supports inference on large,
variable-size tiles (\Cref{sec:inference}); populating
these fields would extend the same large-tile design to the TME-extraction and
narrative stages, so that each large tile's narrative could be indexed and aggregated
toward a slide-level description. This aggregation step is not covered in the present
manuscript.
\texttt{RAW\_FEATURES} contains the
22 numeric outputs of the TME pipeline (\Cref{sec:tme}); \texttt{SCENARIO}
contains the six categorical labels from the rule-based translator;
\texttt{KEYWORDS} is the flat token list. The model is trained to generate the
target clinical narrative autoregressively after this prompt. This four-block
structure separates raw measurements, categorical phenotypes, and vocabulary tokens
into distinct addressable fields while keeping the entire context within one sequence
(average prompt length $\approx$74 words; max sequence length 1,024 tokens).

\paragraph{Training objective.}
The SFT objective minimises the standard cross-entropy loss over the target narrative
token sequence $w_1,\ldots,w_T$, conditioned on the input prompt $\mathbf{x}$:
\begin{equation}
  \mathcal{L}_\text{SFT}(\theta) = -\frac{1}{T}\sum_{t=1}^{T}
  \log P_\theta\!\left(w_t \mid w_1,\ldots,w_{t-1},\,\mathbf{x}\right)
  \label{eq:sft}
\end{equation}
where $P_\theta$ denotes the model's conditional token distribution under parameters
$\theta$. At inference, narrative tokens are generated autoregressively by sampling
from $P_\theta(w_t \mid w_{<t},\mathbf{x})$ until an end-of-sequence token is produced.

The key advantage of using structured JSON as input is that
the model is grounded in measured values, as compared to raw image pixels as input. By taking numerical numbers from structured JSON, every claim in the output narrative can be
traced to a specific quantitative feature (\eg\ ``heavy lymphocytic infiltrate'' maps to
$r_\text{inf}>0.30$; ``immune cells spatially excluded'' maps to
$\bar{d}_{\text{inf}\to\text{neo}}>60$\,px). This grounding substantially reduces
hallucination risk compared to image-conditioned generation, where the model must infer
quantities from visual patterns without explicit access to the underlying numerical
measurements. A representative end-to-end example, including the full numeric feature
set, six categorical interpretation labels, eleven biological tokens, and the
corresponding generated narrative, is shown in \Cref{tab:segtme_output}.

\paragraph{Fine-tuning configuration.}
\Cref{tab:bionemo_sft} summarises the SFT configuration.
The base model is \texttt{meta-llama/Llama-3.2-1B} (1.24\,B parameters,
decoder-only Transformer), pre-trained by Meta on general-domain text.
Fine-tuning is performed via NVIDIA BioNeMo's \texttt{llama3\_native\_te} recipe using
the \texttt{NVLlamaForCausalLM} architecture with TransformerEngine fused kernels,
on paired examples of the form (structured text prompt, target narrative),
with cross-entropy loss computed only on the narrative completion tokens
(prompt tokens masked, label $= -100$).

\begin{table}[htbp]
\centering
\caption{LLaMA-3.2-1B supervised fine-tuning (SFT) configuration via NVIDIA BioNeMo
\texttt{llama3\_native\_te} recipe. Checkpoint at
\texttt{nvidia-bionemo-segtme/bionemo\_tme\_model\_v2/train\_ddp/final\_model/}.
BioNeMo SFT was conducted on a separate single NVIDIA L4 24\,GB GPU distinct from
the 8$\times$A100 allocation used for segmentation training
(\Cref{tab:hparams}), reflecting the substantially smaller compute
footprint of fine-tuning a 1.24\,B-parameter language model relative to the
dual-head ViT-Giant segmentation backbone.}
\label{tab:bionemo_sft}
\small
\begin{tabular}{ll}
\toprule
Parameter & Value \\
\midrule
Base model & \texttt{meta-llama/Llama-3.2-1B} (\texttt{NVLlamaForCausalLM}, $\sim$1.24\,B params) \\
Training framework & NVIDIA BioNeMo \texttt{llama3\_native\_te} + TransformerEngine 2.8.0 \\
Precision & BF16 throughout \\
Input format & Structured text prompt (METADATA $|$ RAW\_FEATURES $|$ SCENARIO $|$ KEYWORDS) \\
Output format & Free-text TME narrative (autoregressive completion) \\
Training examples & 1.6M (\texttt{bionemo\_sft\_jsonl/}, 6 JSONL shards) \\
Training steps & 50,000 \\
Micro batch size & 4 (gradient accumulation: 8 steps; effective batch: 32 sequences) \\
Max sequence length & 1,024 tokens \\
Optimizer & AdamW ($\beta_1=0.9$, $\beta_2=0.95$, weight decay 0.01) \\
Peak learning rate & $2\times10^{-5}$ \\
LR schedule & Linear warmup (1,000 steps) then cosine decay \\
Gradient clipping & Max norm 1.0 \\
Loss masking & Cross-entropy applied to narrative tokens only (prompt tokens masked) \\
Compute & 1$\times$ NVIDIA L4 24\,GB, $\sim$48.5\,h total \\
Initial / final loss & 2.42 / $\approx$0.052 (cross-entropy on narrative tokens) \\
\bottomrule
\end{tabular}
\end{table}

\paragraph{Narrative quality evaluation.}
The BioNeMo pipeline generates clinically plausible TME narratives (pending
formal evaluation).
Quantitative evaluation of narrative fidelity, including BLEU-4, ROUGE-L,
Cohen's $\kappa$ (inter-pathologist agreement on two independent evaluators), and
hallucination rate (proportion of narrative claims unsupported by the input JSON
values), is deferred to the companion paper, where a
full clinical evaluation is conducted. For the current paper, the representative
example in \Cref{tab:segtme_output} and the qualitative NSCLC case study
(Section~\ref{sec:case_study}) serve as qualitative demonstrations of clinical
plausibility.

\FloatBarrier

\begin{table}[htbp]
    \centering
    \caption{\textbf{Representative \fw{} output summaries for a single H\&E patch across different analysis stages.}
    \textbf{(A)} Instance segmentation metrics: total cell count (4,501), per-class
    counts and population fractions, mean nuclear areas, Tumour-Stroma fraction (94.78\%),
    Immune Infiltration fraction (70.83\%), and spatial interaction metrics
    (mean immune-to-tumour distance: 136.14\,px; engaged immune cells: 257).
    \textbf{(B)} Patch spatial context: six categorical phenotype labels and 11 biological
    tokens generated by rule-based translation.
    \textbf{(C)} BioNeMo-fine-tuned LLaMA-3.2-1B generated TME narrative integrating all quantitative findings
    into a structured clinical description.}
    \small
    \renewcommand{\arraystretch}{1.2}
    \begin{tabularx}{\textwidth}{@{} >{\raggedright\arraybackslash\bfseries}p{4cm} X @{}}
        \toprule
        \textbf{Stage} & \textbf{Output of SegTME-UNI2} \\
        \midrule
        
        A. Instance segmentation & 
        \begin{tabular}[t]{@{} p{0.5\linewidth} p{0.45\linewidth} @{}}
            \textbf{Total Cell Count:} 4,501 & \textbf{Mean Spatial Entropy:} 0.2678 \\
            \textbf{Neoplastic Count:} 1,290 & \textbf{Neoplastic Ratio:} 28.66\% \\
            \textbf{Neoplastic Mean Area:} 3,216.37 & \textbf{Immune Count:} 3,133 \\
            \textbf{Immune Ratio:} 69.61\% & \textbf{Immune Mean Area:} 1,970.64 \\
            \textbf{Tumour-Stroma Fraction:} 94.78\% & \textbf{Immune Infiltration Fraction:} 70.83\% \\
            \textbf{Stroma Count:} 71 & \textbf{Stroma Ratio:} 1.58\% \\
            \textbf{Stroma Mean Area:} 2,133.23 & \textbf{Mean Dist (Immune to Tumour):} 136.14 px \\
            \textbf{Min Dist (Immune to Tumour):} 30.94 px & \textbf{Engaged Immune Cells:} 257 \\
            \textbf{Dead Cell Count (Ratio):} 7 (0.16\%) & \textbf{Epithelial Cell Count:} 0 \\
        \end{tabular} \\
        
        \midrule
        
        B. Patch spatial context & 
        \begin{tabular}[t]{@{} >{\bfseries}p{0.4\linewidth} p{0.55\linewidth} @{}}
        	  \textbf{Feature Category:} & \textbf{Interpretation Label:} \\
            Composition Phenotype & Low Tumour Purity \\
            Immune Status & Immune Hot \\
            Stromal Status & Stroma-Free \\
            Spatial Interaction & Excluded \\
            Architecture & Segregated \\
            Morphology & Large/Pleomorphic \\
        \end{tabular} \vspace{0.3cm} \newline
        \textbf{Translation for token:} \vspace{0.1cm} \newline
        \begin{tabular}[t]{@{} p{0.32\linewidth} p{0.32\linewidth} p{0.32\linewidth} @{}}
            $\bullet$ Immune-dominant & $\bullet$ Lack of fibrosis & $\bullet$ Homogeneous \\
            $\bullet$ Lymphocyte-rich & $\bullet$ Immune exclusion & $\bullet$ Nuclear enlargement \\
            $\bullet$ Heavy infiltrate & $\bullet$ Restricted to stroma & $\bullet$ Anisocytosis \\
            $\bullet$ Non-desmoplastic & $\bullet$ Bystander immune cells & \\
        \end{tabular} \\
        
        \midrule
        
        C. Patch-to-text narratives & 
        The section exhibits low tumour purity (Neoplastic ratio: 0.29), consisting primarily of non-neoplastic elements. Cytologically, the neoplastic cells display enlarged, pleomorphic features suggestive of nuclear atypia. Notably, there is an absence of significant desmoplastic reaction or fibrous stroma (Connective ratio: 0.02). Although presenting a heavy lymphocytic infiltrate ('Immune Hot'), the immune cells are spatially excluded from the tumour core and restricted to the surrounding tissue (Mean distance: 136.1px). However, the majority of the immune cells appear to be bystanders with minimal direct tumour engagement. The spatial architecture is highly homogeneous and segregated (Entropy: 0.27). \\
        
        \bottomrule
    \end{tabularx}
    \label{tab:segtme_output}
\end{table}

\subsection{Inference and Post-Processing}
\label{sec:inference}

Whole-slide deployment is handled by \texttt{LargeTilePredictor}, a standalone,
installable component (\texttt{pip install segtme-uni2};
\texttt{from segtme import UNI2UperHoVer, LargeTilePredictor}) that accepts a
whole-slide image (WSI) and a target tile size (not one fixed patch size). It
places a grid of large tiles across the slide with a stride, the physical spacing
between successive tile positions, that scales with how much tissue each tile
covers: a tile at TCGA-UT's coarser resolution covers substantially more tissue per
placement than one at PanNuke's native resolution would, so covering the same slide
needs a correspondingly larger stride and far fewer tile placements. This is the
practical payoff of adapting \seg{} to TCGA-UT (\Cref{sec:tcga_dataset}): the same
backbone that gains the spatial context TME characterisation requires also becomes
cheaper to run across a whole slide. Each large tile is then processed independently
by one of the two regimes below, and its resulting TME features and narrative
(\Cref{sec:tme,sec:bionemo}) are returned per tile; aggregating these into a single
slide-level output is not implemented in the present release
(\Cref{sec:bionemo}).

\begin{figure}[t]
    \centering
    \includegraphics[width=0.6\textwidth]{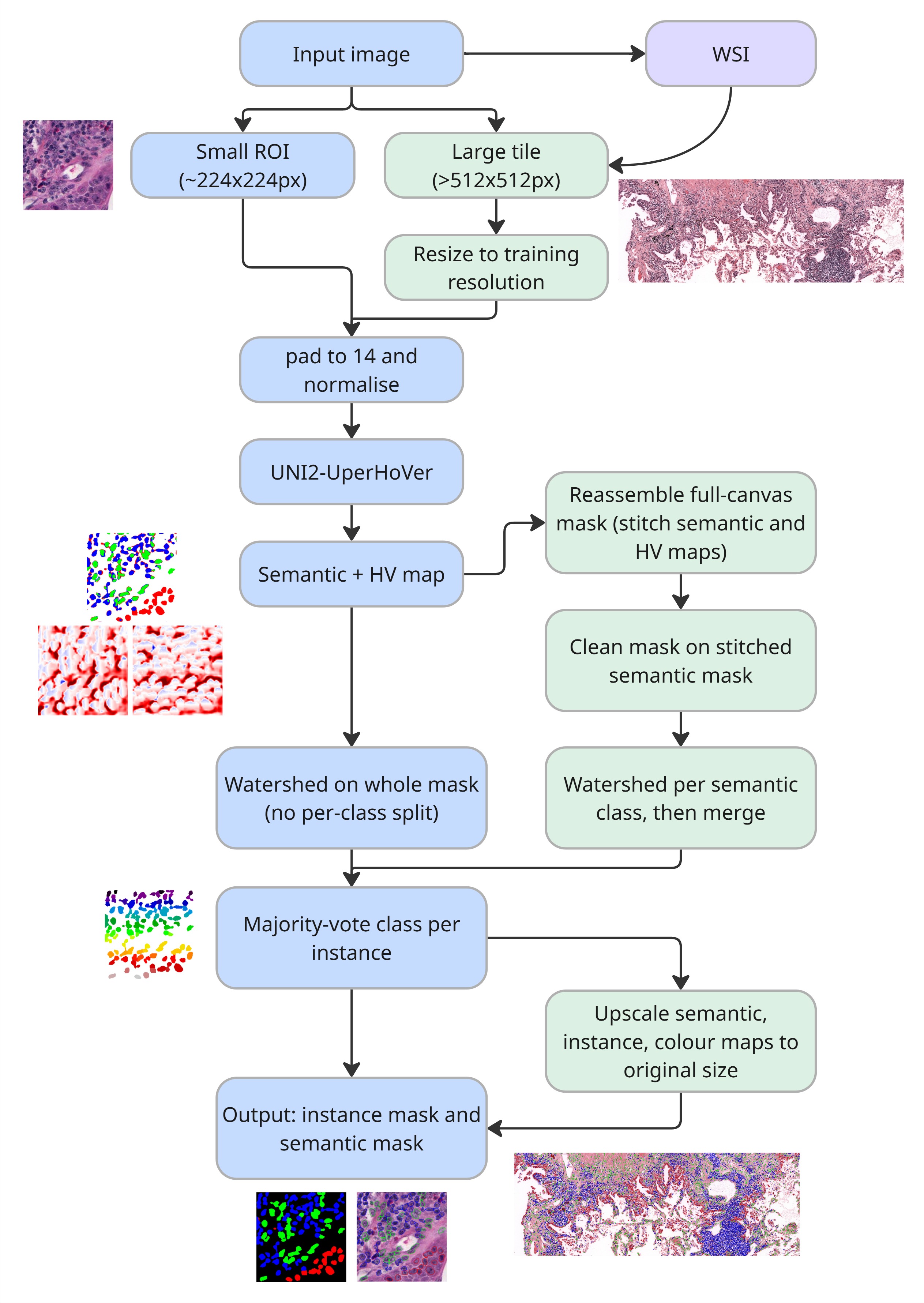}
    \caption{Overview of the inference and post-processing pipeline described in
    \Cref{sec:inference}, shown for both the single-tile regime (blue; small
    ROI, $\sim$224$\times$224\,px) and the whole-slide/tiled regime (green; large
    tiles extracted from a WSI). Large tiles are first resized to the canonical
    training resolution, while small ROIs are used as-is; both are then padded to
    the nearest multiple of 14\,px, normalised, and passed through the shared
    UNI2-UperHoVer backbone-decoder to produce a semantic map and a two-channel
    horizontal-vertical (HV) map. In the whole-slide regime, overlapping tile
    predictions are stitched into a full-resolution canvas and refined with a
    morphological opening before watershed is run independently per semantic class
    and the resulting instances merged; in the single-tile regime, no stitching or
    cleaning is required, and watershed is instead applied once over the entire
    mask. Both regimes converge at a majority-vote step that assigns each instance
    a class label from the semantic map. Whole-slide outputs are then upscaled to
    the original image resolution via nearest-neighbour interpolation, whereas
    single-tile outputs, already at native resolution, require no rescaling; both
    terminate in the same instance-mask and semantic-mask output. Representative
    inputs and intermediate/final outputs (H\&E crop, decoded semantic map, HV
    maps, colour-coded instance map, and class overlay) are shown alongside their
    corresponding pipeline stage.}
    \label{fig:inference_pipeline}
\end{figure}

Two complementary inference regimes are used depending on the scale of the input,
sharing the same backbone-decoder network, pre-processing, and instance-decoding logic
described below, and summarised in \Cref{fig:inference_pipeline}.

For regions of interest that are already at, or close to, the target resolution of approximately $224\times224$ pixels, the crop is
treated as a single tile: it is zero-padded to the nearest multiple of 14 pixels,
satisfying the ViT-Giant patch-size constraint, normalised, and passed through the
backbone-decoder pair in a single forward pass to produce a semantic logit map and a
two-channel horizontal-vertical (HV) distance map. No downscaling, tiling, stitching, or
morphological cleaning is required in this regime, since the tile already constitutes the
full prediction canvas.

For full-resolution images larger than $512\times512$ pixels, whole-image inference
at native resolution is computationally intractable for a ViT backbone trained at
$224\times224$ pixels, given the quadratic memory cost of self-attention in patch-token
sequence length; we therefore use MPP-normalised tiled processing. This $512\times512$-pixel
threshold is the native-resolution crop size that, at the dataset's native
$0.5\,\mu\text{m/pixel}$ scanning resolution, corresponds to Scale~5's coarsest effective
resolution ($1.0\,\mu\text{m/pixel}$) once downsampled to the standard
$256\times256$-pixel patch size (\Cref{sec:tcga_dataset}). The large input image
is downscaled by $s = \text{MPP}_\text{input} / 0.314$ to a canonical resolution of
approximately $0.35\,\mu\text{m/pixel}$, then partitioned into $224\times224$-pixel tiles
with a stride of 112\,px (50\% overlap). Each tile is zero-padded and normalised as above,
then passed through the same backbone-decoder pair. Semantic maps from overlapping tiles
are stitched using foreground-priority blending: each pixel adopts the foreground class
prediction from any overlapping tile that assigns it a non-background class; HV maps use
last-writer overwrite. The stitched semantic mask is then refined with a morphological
opening to suppress noise introduced at tile boundaries before instance separation.

Instance segmentation applies a marker-controlled watershed to the HV energy surface:
seed markers are identified as local maxima of the Euclidean distance transform (minimum
separation 7\,px); watershed is run with compactness 0.01 on the combined Sobel-HV
energy map; objects smaller than 30\,px$^2$ are removed as spurious detections. For
whole-slide inference, this procedure is applied independently within each predicted
semantic class before merging the resulting instance labels, preventing adjacent nuclei
of different classes from being fused into a single instance across tile boundaries; for
single-tile inference, where no such cross-tile bleeding can occur, watershed is instead
applied once over the entire mask. Instance class labels are assigned by majority vote of
the semantic segmentation map within each instance region, preserving consistency between
the semantic and instance outputs without requiring any additional inference pass.
Whole-slide outputs are finally upscaled to the original image resolution via
nearest-neighbour interpolation; single-tile outputs, already at native resolution,
require no such rescaling.

\FloatBarrier

\section{Results}
\label{sec:results}

\subsection{Semantic segmentation: internal validation}

\Cref{tab:seg_results} reports evaluation results across two protocols.
Part~(A) evaluates all three curriculum-stage models on the same PanNuke 20\%
held-out test set, which contains human-annotated ground truth, enabling fair
cross-stage comparison on validated labels. Because $\mathcal{M}_2$ and
$\mathcal{M}_3$ are trained on TCGA-UT pseudo-labels (not on PanNuke), Part~(A)
is a \textbf{cross-domain evaluation} for these two models: it assesses whether
pseudo-label training preserves the cell-type taxonomy learned from PanNuke, not
in-domain accuracy. Lower Part~(A) scores for $\mathcal{M}_2$ and $\mathcal{M}_3$
therefore reflect \emph{domain shift away from PanNuke}, not model degradation.
Part~(B) reports self-consistency on each model's own TCGA-UT held-out pseudo-label
partition; these values indicate training convergence but cannot be interpreted as
accuracy against human ground truth. The baseline is a single-head UperNet
(semantic only, no HV head) with the same \bb{} backbone and training configuration
as \seg{}, trained on PanNuke.

The curriculum's success criterion is therefore (i)~increasing pseudo-label
self-consistency in Part~(B), and (ii)~the training dynamics in
\Cref{fig:training_curves}, which show each stage's initial mIoU (0.41 to 0.47 to
0.53) progressively higher than the preceding stage. This is a direct evidence that each
stage generates better pseudo-labels than the last. $\mathcal{M}_3$'s Part~(A)
collapse to 0.1587 is additionally attributable to a systematic magnification
mismatch detailed in Section~\ref{sec:discussion}, most pronounced on Scales~3
through~5; a corrected Stage~3 retraining is planned as the next course of action.

\begin{figure}[htbp]
\centering
\begin{tabular}{lccccc}
Original & \includegraphics[width=0.14\textwidth]{figures/image_500.png} &
\includegraphics[width=0.14\textwidth]{figures/image_1000.png} &
\includegraphics[width=0.14\textwidth]{figures/image_1500.png} &
\includegraphics[width=0.14\textwidth]{figures/image_2000.png} &
\includegraphics[width=0.14\textwidth]{figures/image_2500.png} \\
PanNuke GT & \includegraphics[width=0.14\textwidth]{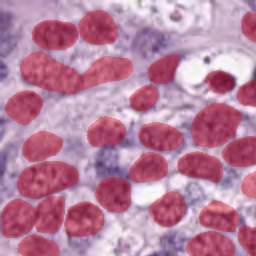} &
\includegraphics[width=0.14\textwidth]{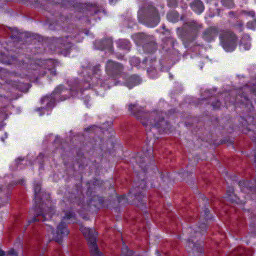} &
\includegraphics[width=0.14\textwidth]{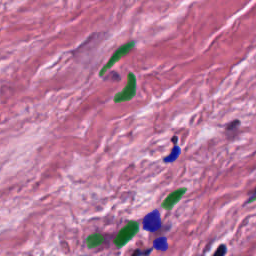} &
\includegraphics[width=0.14\textwidth]{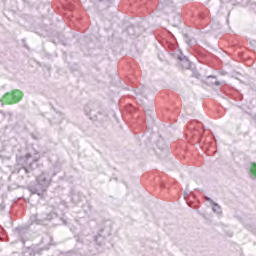} &
\includegraphics[width=0.14\textwidth]{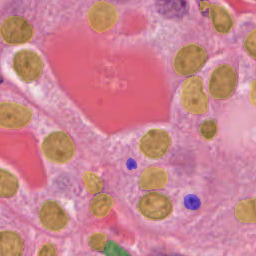} \\
Model $\mathcal{M}_1$ & \includegraphics[width=0.14\textwidth]{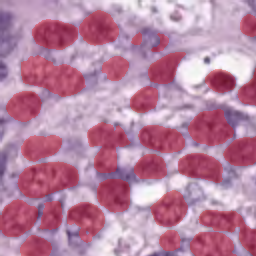} &
\includegraphics[width=0.14\textwidth]{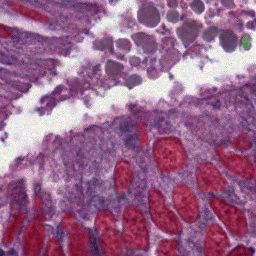} &
\includegraphics[width=0.14\textwidth]{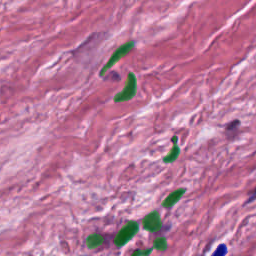} &
\includegraphics[width=0.14\textwidth]{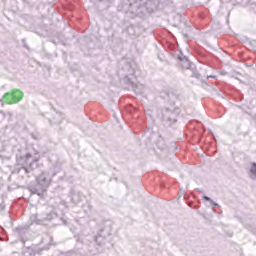} &
\includegraphics[width=0.14\textwidth]{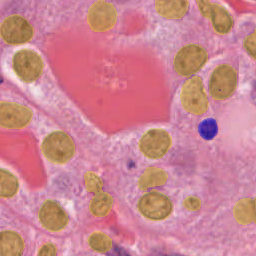} \\
Model $\mathcal{M}_2$ & \includegraphics[width=0.14\textwidth]{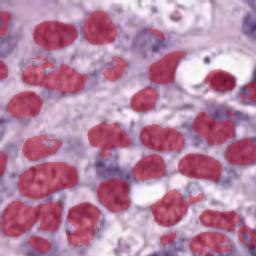} &
\includegraphics[width=0.14\textwidth]{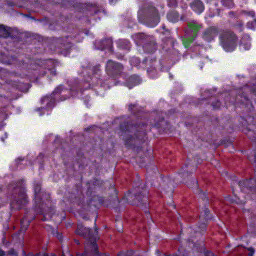} &
\includegraphics[width=0.14\textwidth]{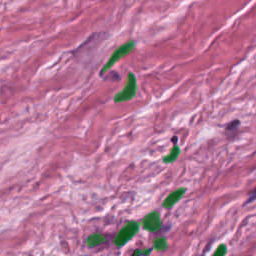} &
\includegraphics[width=0.14\textwidth]{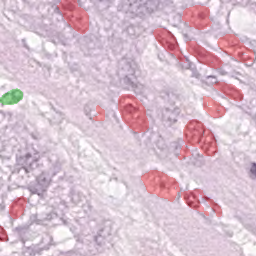} &
\includegraphics[width=0.14\textwidth]{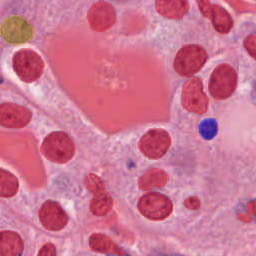} \\
Model $\mathcal{M}_3$ & \includegraphics[width=0.14\textwidth]{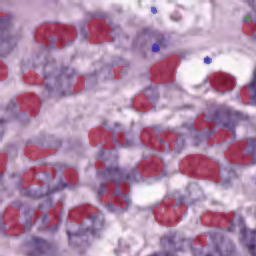} &
\includegraphics[width=0.14\textwidth]{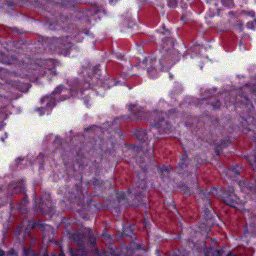} &
\includegraphics[width=0.14\textwidth]{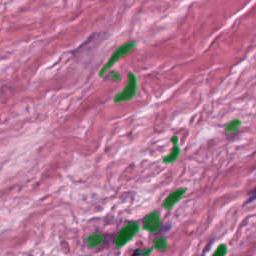} &
\includegraphics[width=0.14\textwidth]{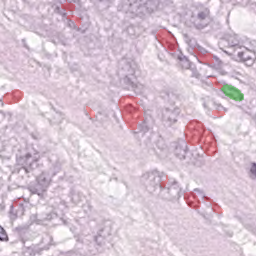} &
\includegraphics[width=0.14\textwidth]{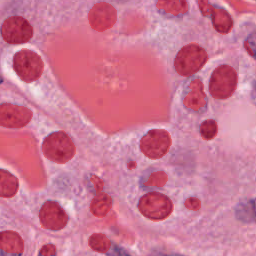} 
\end{tabular}
\caption{\textbf{Semantic segmentation result on representative samples of PanNuke dataset - fold 3.}
Columns show the same five representative patches used in \Cref{fig:pannuke_samples} (Image 500, 1000, 1500, 2000, 2500). Rows show, top to bottom: the original H\&E patch, the PanNuke ground-truth six-class semantic mask, and predictions from $\mathcal{M}_1$ (Stage~1, PanNuke-supervised), $\mathcal{M}_2$ (Stage~2, TCGA-UT Scale~0 pseudo-labels), and $\mathcal{M}_3$ (Stage~3, TCGA-UT Scales~0--5 pseudo-labels), all evaluated against this human-annotated PanNuke ground truth. The visible qualitative degradation from $\mathcal{M}_1$ to $\mathcal{M}_3$ is consistent with the quantitative PanNuke-evaluated mIoU collapse reported in Part~(A) of \Cref{tab:seg_results}.}
\label{fig:pannuke_results}
\end{figure}

The Dead class consistently produces the lowest IoU across all models
($\mathcal{M}_1$: 0.1345), reflecting its inherent difficulty: dead and apoptotic
nuclei are morphologically heterogeneous (pyknotic, karyorrhectic, and ghost nuclei
all map to the same class label), numerically rare in PanNuke (representing $<3\%$ of
all annotated nuclei), and frequently confused with small Inflammatory cells due to
nuclear size overlap. This class-specific difficulty is consistent with the original
PanNuke benchmark report~\cite{gamper2020pannuke} and is not specific to \seg{}. Example of the semantic segmentation overlay using all three models are shown in \Cref{fig:pannuke_results}.

\FloatBarrier

\FloatBarrier

\begin{table}[htbp]
\centering
\caption{Semantic segmentation evaluation results (preliminary; single training run per stage,
no multi-seed variance reported.
Part~(A) evaluates all models on the PanNuke 20\% held-out test set (human-annotated
ground truth, GT), enabling direct cross-stage comparison on validated labels.
Part~(B) reports pseudo-label self-consistency on each model's respective TCGA-UT
held-out partition; these figures reflect convergence on pseudo-labelled data and
should not be interpreted as absolute accuracy against human ground truth (PL-val).
mIoU: macro-averaged IoU across five foreground classes (Background excluded); Neo: Neoplastic; Inf: Inflammatory; Conn: Connective; Dead: Dead cells; Epi: Epithelial.
$^\dagger$Cross-domain: model trained on TCGA-UT pseudo-labels, evaluated on PanNuke
ground truth; lower scores reflect the domain shift (see Section~\ref{sec:discussion} for Stage~3 magnification mismatch diagnosis).
$^\ddagger$$\mathcal{M}_1$ evaluated on TCGA-UT s0 PL-val: mIoU 0.3631 (5-class, BG excluded),
confirming that $\mathcal{M}_2$ (0.3666) outperforms $\mathcal{M}_1$ on the TCGA-UT
domain after Stage~2 pseudo-label training.}
\label{tab:seg_results}
\small
\begin{tabularx}{\textwidth}{llccccccc}
\toprule
\multirow{2}{*}{Model} &
\multirow{2}{*}{Eval set} &
\multirow{2}{*}{mIoU $\uparrow$} &
\multicolumn{5}{c}{Per-class IoU ($\uparrow$)} \\
\cmidrule(lr){4-8}
 & & & Neo & Inf & Conn & Dead & Epi \\
\midrule
\multicolumn{8}{l}{\textit{(A) Human-labelled evaluation: PanNuke 20\% held-out (GT)}}\\
\midrule
UperNet single-head (baseline)
  & PanNuke (GT)
  & 0.4904
  & 0.6625 & 0.4555
  & 0.5384 & 0.1140 & 0.6819 \\
$\mathcal{M}_1$ (Stage 1, PanNuke)
  & PanNuke (GT)
  & 0.5039
  & 0.6675 & 0.4764
  & 0.5562 & 0.1345 & 0.6849 \\
$\mathcal{M}_2$ (Stage 2, TCGA-UT s0)$^\dagger$
  & PanNuke (GT)
  & 0.3292
  & 0.5446 & 0.3552
  & 0.4246 & 0.0660 & 0.2553 \\
$\mathcal{M}_3$ (Stage 3, TCGA-UT s0--5)$^\dagger$
  & PanNuke (GT)
  & 0.1587
  & 0.4026 & 0.0880
  & 0.2897 & 0.0078 & 0.0051 \\
\midrule
\multicolumn{8}{l}{\textit{(B) Pseudo-label self-consistency: TCGA-UT 20\% held-out (PL-val, not human-annotated)}}\\
\midrule
$\mathcal{M}_1$ (Stage 1, PanNuke)$^\ddagger$
  & TCGA-UT s0 (PL-val)
  & 0.3631
  & 0.5708 & 0.3527
  & 0.3776 & 0.1278 & 0.3863 \\
$\mathcal{M}_2$ (Stage 2)
  & TCGA-UT s0 (PL-val)
  & 0.3666
  & 0.5862 & 0.3624
  & 0.3953 & 0.0884 & 0.4008 \\
$\mathcal{M}_3$ (Stage 3)
  & TCGA-UT s0--5 (PL-val)
  & 0.4425
  & 0.6824 & 0.4620
  & 0.4663 & 0.1788 & 0.4231 \\
\bottomrule
\end{tabularx}
\end{table}

\FloatBarrier

\subsection{Training dynamics} \label{sec:training_dynamics}

\Cref{fig:training_curves} shows the validation mIoU and training loss trajectories
for all three curriculum stages. All three models initialise from the same \bb{}
pretrained backbone with randomly initialised decoder heads; the progressive improvement
observed across stages reflects increasing pseudo-label quality, where the weight is not inherited between stages.

\paragraph{Stage 1 ($\mathcal{M}_1$, PanNuke).}
$\mathcal{M}_1$ begins at a validation mIoU of 0.41 at the first checkpoint and
improves rapidly during the first 25\% of training, reaching 0.83 mIoU as the model
learns the six-class taxonomy from clean human annotations. The final training loss of 0.025 reflects the
well-constrained nature of the PanNuke training set.

\paragraph{Stage 2 ($\mathcal{M}_2$, TCGA-UT Scale 0).}
$\mathcal{M}_2$ initialises fresh from \bb{} and begins training on Scale-0 TCGA-UT
pseudo-labels generated by $\mathcal{M}_1$. The initial mIoU of 0.47 is higher than
$\mathcal{M}_1$'s starting point (0.41), reflecting the higher quality of
pseudo-label initialisation compared to the random decoder alone. The final training loss of 0.061 is higher than
Stage~1, consistent with the reduced label quality of pseudo-annotated data.

\paragraph{Stage 3 ($\mathcal{M}_3$, TCGA-UT Scales 0--5).}
$\mathcal{M}_3$ trains on the full 1.6M-patch multi-resolution corpus using
pseudo-labels generated by $\mathcal{M}_2$. Training runs for 100 epochs; because each
epoch processes all 1.6M patches, Stage~3 accumulates substantially more total
gradient steps (335,100) than Stage~1 (24,750) or Stage~2 (212,500) despite the lower epoch count. 
The higher final training loss of 0.120 reflects the greater label noise and domain
diversity introduced by five additional resolution scales not seen in Stages~1 or~2.
Notably, the mIoU curve for $\mathcal{M}_3$ continues to improve steadily at the
100-epoch cutoff, suggesting that further training would yield additional gains.

\begin{figure}[htbp]
\centering
\begin{tabular}{lcccccc}
   & \includegraphics[width=0.14\textwidth]{figures/0-TCGA-BA-5558-01Z-00-DX1_0_7_520.jpg} &
\includegraphics[width=0.14\textwidth]{figures/1-TCGA-BA-5558-01Z-00-DX1_0_2_624.jpg} &
\includegraphics[width=0.14\textwidth]{figures/2-TCGA-BA-5558-01Z-00-DX1_0_1_728.jpg} &
\includegraphics[width=0.14\textwidth]{figures/3-TCGA-BA-5558-01Z-00-DX1_0_3_832.jpg} &
\includegraphics[width=0.14\textwidth]{figures/4-TCGA-BA-5558-01Z-00-DX1_0_3_936.jpg} &
\includegraphics[width=0.14\textwidth]{figures/5-TCGA-BA-5558-01Z-00-DX1_0_5_1041.jpg} \\
$\mathcal{M}_1$ & \includegraphics[width=0.14\textwidth]{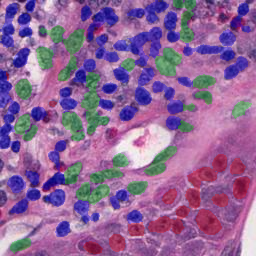} &
\includegraphics[width=0.14\textwidth]{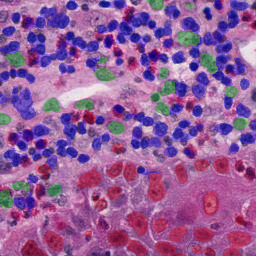} &
\includegraphics[width=0.14\textwidth]{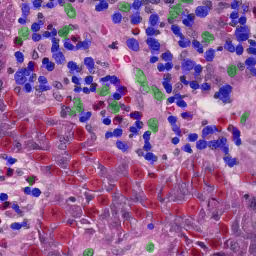} &
\includegraphics[width=0.14\textwidth]{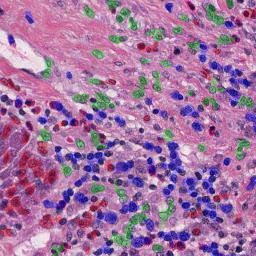} &
\includegraphics[width=0.14\textwidth]{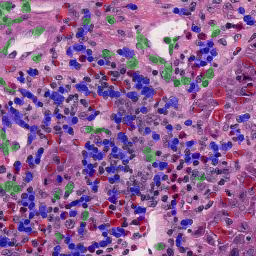} &
\includegraphics[width=0.14\textwidth]{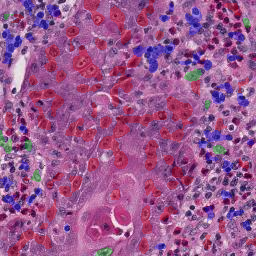} \\
$\mathcal{M}_2$ & \includegraphics[width=0.14\textwidth]{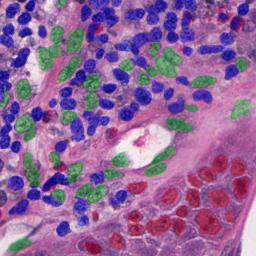} &
\includegraphics[width=0.14\textwidth]{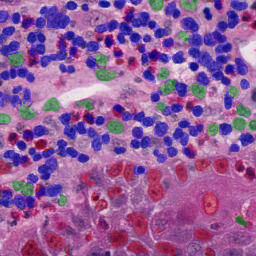} &
\includegraphics[width=0.14\textwidth]{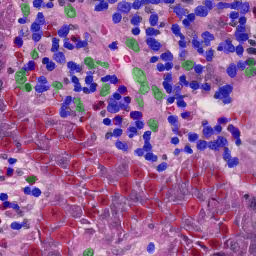} &
\includegraphics[width=0.14\textwidth]{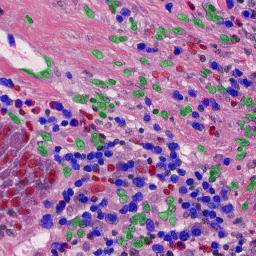} &
\includegraphics[width=0.14\textwidth]{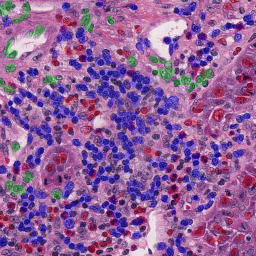} &
\includegraphics[width=0.14\textwidth]{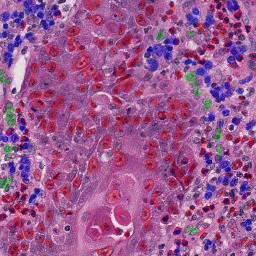} \\
$\mathcal{M}_3$ & \includegraphics[width=0.14\textwidth]{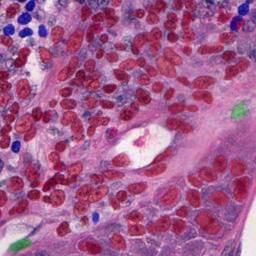} &
\includegraphics[width=0.14\textwidth]{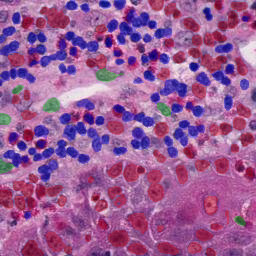} &
\includegraphics[width=0.14\textwidth]{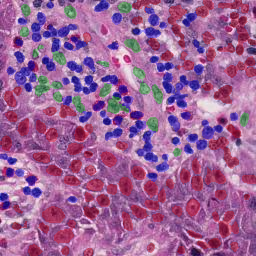} &
\includegraphics[width=0.14\textwidth]{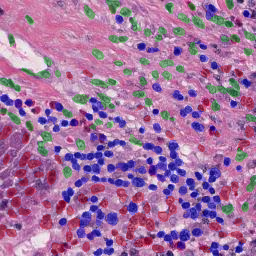} &
\includegraphics[width=0.14\textwidth]{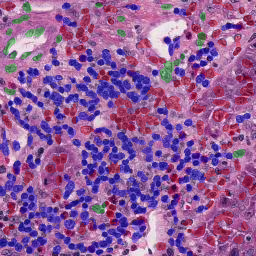} &
\includegraphics[width=0.14\textwidth]{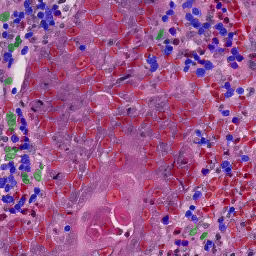} 
\end{tabular}
\caption{\textbf{Semantic segmentation result on TCGA-UT samples from tissue region TCGA-BA-5558, Scale~0 to Scale~5.}
Columns correspond to the same tissue region shown in \Cref{fig:tcga_scales}, sampled at each of the six TCGA-UT resolution scales (Scale~0, 0.5\,$\mu$m/pixel, through Scale~5, 1.0\,$\mu$m/pixel). Rows show, top to bottom: the original H\&E patch and the predicted six-class semantic segmentation from $\mathcal{M}_1$, $\mathcal{M}_2$, and $\mathcal{M}_3$. No human-annotated ground truth exists for TCGA-UT, so this figure is qualitative only. Visible degradation in $\mathcal{M}_3$'s predictions at the coarser scales (Scale~3 to Scale~5) illustrates the scale-normalisation mismatch diagnosed in \Cref{sec:discussion}.}

\label{fig:tcga_scale_results}
\end{figure}

\paragraph{Cross-stage comparison.}
The most direct evidence that the progressive curriculum succeeds is
\Cref{fig:training_curves}(a): each model's \emph{initial} validation mIoU
(at training step~0, before any gradient updates on the new corpus) is higher than
the preceding stage's: 0.41 ($\mathcal{M}_1$, random decoder), 0.47 ($\mathcal{M}_2$),
0.53 ($\mathcal{M}_3$), reflecting the compounding quality of pseudo-labels across
stages. A static decoder randomly initialised from \bb{} achieves 0.41 mIoU on PanNuke
at step~0; the same decoder exposed to $\mathcal{M}_2$ pseudo-labels from day one starts
at 0.47. This is the \emph{central empirical finding} of the paper: the progressive
curriculum measurably improves TME characterisation without any additional human
annotation at each stage. \Cref{fig:tcga_scale_results} complements this quantitative
trend with a qualitative view on TCGA-UT itself: all three models produce visually
coherent six-class segmentation at Scale~0, but $\mathcal{M}_3$'s predictions visibly
degrade at the coarser scales (Scale~3 to Scale~5), consistent with the
scale-normalisation mismatch. $\mathcal{M}_3$, despite training on a broader six-scale TCGA-UT distribution,
currently yields lower segmentation quality owing to a systematic magnification
mismatch in its Stage~3 pseudo-label targets, which is an internal training artefact
unrelated to the resolution of the input patches, as characterised in
Section~\ref{sec:discussion}. Further analysis will be using $\mathcal{M}_2$ checkpoint as qualitative evaluation.

\FloatBarrier

\begin{figure}[htbp]
\centering
\resizebox{\textwidth}{!}{%
\begin{tikzpicture}
\begin{groupplot}[
    group style={group size=2 by 1, horizontal sep=2.2cm},
    width=7.5cm, height=6cm,
    grid=major, grid style={line width=0.3pt, draw=gray!25},
    tick align=outside,
    ticklabel style={font=\scriptsize},
    enlarge x limits=false,
]
\nextgroupplot[
    title={\textbf{(a)\ Validation mIoU}},
    xlabel={Training Progress (\%)},
    ylabel={Mean IoU},
    xmin=0, xmax=105,
    ymin=0.35, ymax=0.97,
    ytick={0.4,0.5,0.6,0.7,0.8,0.9},
    xtick={0,25,50,75,100},
    legend pos=south east,
    legend style={font=\scriptsize, fill=white, draw=gray!40,
                  inner sep=3pt, row sep=0.5pt},
    legend cell align=left,
]
\addplot[color=s1col, line width=1.3pt, smooth] coordinates {
    (0.40,0.4077)(5.22,0.7131)(10.04,0.7839)(14.86,0.7982)(19.68,0.8009)
    (24.50,0.8293)(29.32,0.8370)(34.14,0.8508)(38.96,0.8612)(43.78,0.8701)
    (48.59,0.8596)(53.41,0.8895)(58.23,0.8901)(63.05,0.8956)(67.87,0.9007)
    (72.69,0.9114)(77.51,0.9122)(82.33,0.9170)(87.15,0.9220)(91.97,0.9268)
    (96.79,0.9300)(100.00,0.9313)};
\addlegendentry{$\mathcal{M}_1$ (PanNuke)};
\addplot[color=s1col, only marks, mark=star, mark size=5.5pt,
         mark options={fill=s1col}, forget plot] coordinates {(100.00,0.9313)};
\addplot[color=s2col, line width=1.3pt, smooth] coordinates {
    (0.24,0.4714)(5.65,0.6024)(11.06,0.6646)(16.47,0.6954)(21.88,0.7148)
    (27.29,0.7291)(32.71,0.7430)(38.12,0.7503)(43.53,0.7575)(48.94,0.7682)
    (54.35,0.7734)(59.76,0.7790)(65.18,0.7872)(70.59,0.7934)(76.00,0.8008)
    (81.41,0.8073)(86.82,0.8113)(92.24,0.8158)(97.65,0.8190)(100.00,0.8197)};
\addlegendentry{$\mathcal{M}_2$ (TCGA-UT s0)};
\addplot[color=s2col, only marks, mark=star, mark size=5.5pt,
         mark options={fill=s2col}, forget plot] coordinates {(100.00,0.8197)};
\addplot[color=s3col, line width=1.3pt, smooth] coordinates {
    (1.49,0.5345)(4.48,0.5705)(7.46,0.6007)(10.44,0.6272)(13.43,0.6526)
    (16.41,0.6685)(19.40,0.6822)(22.38,0.6896)(25.37,0.6936)(29.84,0.7015)
    (32.83,0.7115)(35.81,0.7208)(38.79,0.7217)(41.78,0.7299)(44.76,0.7340)
    (47.75,0.7371)(50.73,0.7413)(53.72,0.7451)(56.70,0.7482)(59.68,0.7506)
    (62.67,0.7526)(65.65,0.7552)(68.64,0.7576)(71.62,0.7595)(74.60,0.7614)
    (77.59,0.7629)(80.57,0.7644)(83.56,0.7664)(86.54,0.7673)(89.53,0.7689)
    (92.51,0.7704)(95.49,0.7713)(99.97,0.7724)};
\addlegendentry{$\mathcal{M}_3$ (TCGA-UT s0--5)};
\addplot[color=s3col, only marks, mark=star, mark size=5.5pt,
         mark options={fill=s3col}, forget plot] coordinates {(99.97,0.7724)};
\addlegendimage{only marks, mark=star, mark size=5.5pt, black,
                mark options={fill=black}};
\addlegendentry{Best checkpoint ($\star$)};
\nextgroupplot[
    title={\textbf{(b)\ Training Loss ($\mathcal{L}_{\mathrm{total}}$)}},
    xlabel={Training Progress (\%)},
    ylabel={Loss},
    xmin=0, xmax=105,
    ymode=log, ymin=0.015, ymax=1.5,
    ytick={0.02,0.05,0.10,0.20,0.50,1.00},
    yticklabels={0.02,0.05,0.10,0.20,0.50,1.00},
    xtick={0,25,50,75,100},
    legend pos=north east,
    legend style={font=\scriptsize, fill=white, draw=gray!40,
                  inner sep=3pt, row sep=0.5pt},
    legend cell align=left,
]
\addplot[color=s1col, line width=1.3pt] coordinates {
    (0.20,0.8784)(4.06,0.1273)(7.91,0.0786)(11.76,0.0648)(15.62,0.0594)
    (19.47,0.0571)(23.33,0.0482)(27.18,0.0462)(31.03,0.0455)(34.89,0.0405)
    (38.74,0.0391)(42.59,0.0378)(46.45,0.0355)(50.30,0.0366)(54.16,0.0319)
    (58.01,0.0317)(61.86,0.0307)(65.72,0.0298)(69.57,0.0291)(73.43,0.0273)
    (77.28,0.0270)(81.13,0.0266)(84.99,0.0252)(88.84,0.0249)(92.69,0.0244)
    (96.55,0.0242)(100.00,0.0245)};
\addlegendentry{$\mathcal{M}_1$ (final: 0.025)};
\addplot[color=s2col, line width=1.3pt] coordinates {
    (0.02,0.8661)(4.02,0.2029)(8.02,0.1543)(12.02,0.1316)(16.02,0.1159)
    (20.02,0.1073)(24.02,0.1019)(28.02,0.0969)(32.02,0.0927)(36.02,0.0879)
    (40.02,0.0873)(44.02,0.0837)(48.02,0.0813)(52.02,0.0783)(56.02,0.0766)
    (60.02,0.0752)(64.02,0.0723)(68.02,0.0710)(72.02,0.0694)(76.02,0.0679)
    (80.02,0.0662)(84.02,0.0649)(88.02,0.0632)(92.02,0.0623)(96.02,0.0612)
    (100.00,0.0614)};
\addlegendentry{$\mathcal{M}_2$ (final: 0.061)};
\addplot[color=s3col, line width=1.3pt] coordinates {
    (0.01,0.9962)(4.01,0.1972)(8.01,0.1822)(12.01,0.1711)(16.01,0.1621)
    (20.01,0.1577)(24.01,0.1518)(28.01,0.1528)(32.01,0.1474)(36.00,0.1434)
    (40.00,0.1418)(44.00,0.1409)(48.00,0.1389)(52.00,0.1348)(56.00,0.1331)
    (60.00,0.1308)(64.00,0.1301)(67.99,0.1283)(71.99,0.1273)(75.99,0.1253)
    (79.99,0.1242)(83.99,0.1236)(87.99,0.1229)(91.99,0.1210)(95.99,0.1206)
    (100.00,0.1197)};
\addlegendentry{$\mathcal{M}_3$ (final: 0.120)};
\end{groupplot}
\end{tikzpicture}
}
\caption{\textbf{Training dynamics across the three curriculum stages
($\mathcal{M}_1$: PanNuke, 250 epochs; $\mathcal{M}_2$: TCGA-UT Scale~0, 250 epochs;
$\mathcal{M}_3$: TCGA-UT Scales~0--5, 100 epochs; all seeds identical; single run per stage).}
\textit{Left} (a): validation mIoU trajectories. All three models initialise from
the \bb{} pretrained backbone with randomly initialised decoder heads; no weights are
inherited between stages.
\emph{Key observation:} the initial mIoU at step~0 rises monotonically across stages
(0.41 to 0.47 to 0.53), directly demonstrating that each stage's pseudo-labels
provide a better initialisation than the preceding stage.
\textit{Right} (b): training loss convergence. Final values are
$\mathcal{L}_\text{total}^{(\mathcal{M}_1)} = 0.025$,
$\mathcal{L}_\text{total}^{(\mathcal{M}_2)} = 0.061$,
$\mathcal{L}_\text{total}^{(\mathcal{M}_3)} = 0.120$;
higher values in later stages reflect increased label noise from pseudo-labelling. Stars ($\star$) mark best checkpoints.}
\label{fig:training_curves}
\end{figure}

\FloatBarrier

\FloatBarrier

\subsection{Qualitative case study: TME characterisation in non-small cell lung cancer}
\label{sec:case_study}

To demonstrate the generalisability of \fw{} beyond the TCGA-UT training corpus,
we applied the full inference and TME characterisation pipeline to five H\&E-stained
whole-slide image tiles from the publicly available IGNITE data
toolkit~\cite{lucassen2025ignite}, a multi-centric, multi-scanner non-small cell
lung cancer (NSCLC) histopathology dataset (887 annotated regions of interest from
155 patients).
NSCLC presents particular challenges for automated TME analysis: staining intensity
and protocol vary across the toolkit's multiple contributing centres
and scanners, tissue architecture spans densely cellular tumour nests to fibrous
desmoplastic stroma, and immune infiltration patterns are a known determinant of
immunotherapy response in NSCLC~\cite{lucassen2025ignite}.
These five tiles were selected to represent five distinct and clinically meaningful
TME phenotypes spanning high and low tumour purity, immune-hot and immune-cold
patterns, and histological architecture diversity (high-mixing, segregated, or intermediate/moderately mixed), testing the robustness of \seg{} to the inter-institutional and
inter-scanner stain variability inherent to the IGNITE toolkit's multi-centric design.
All five tiles were processed using $\mathcal{M}_2$ (Stage~2 checkpoint,
trained exclusively on TCGA-UT Scale~0 at 0.5\,$\mu$m/pixel). 

\begin{figure}[htbp]
\centering
\begin{tabular}{>{\centering\arraybackslash}m{0.15\textwidth}
                >{\centering\arraybackslash}m{0.15\textwidth}
                >{\centering\arraybackslash}m{0.15\textwidth}
                >{\centering\arraybackslash}m{0.15\textwidth}
		>{\centering\arraybackslash}m{0.15\textwidth}
}
    {\includegraphics[width=0.15\textwidth]{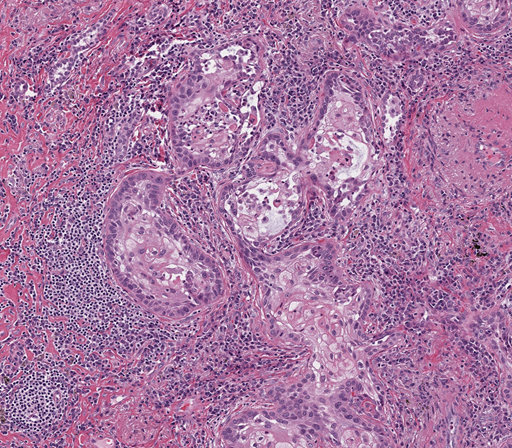}} &
    {\includegraphics[width=0.15\textwidth]{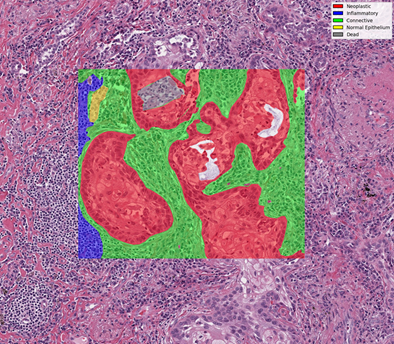}} &
    {\includegraphics[width=0.15\textwidth]{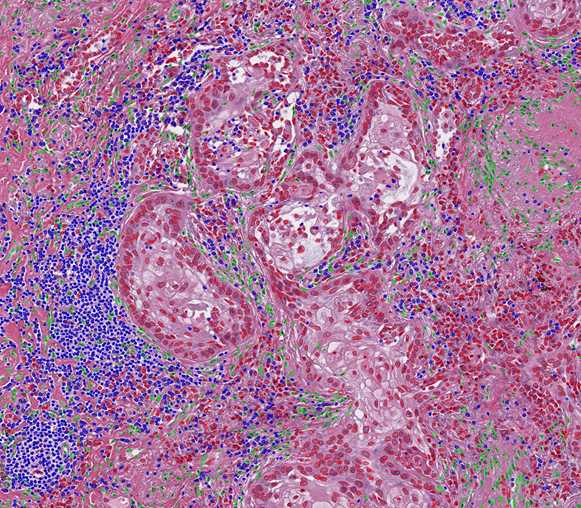}} &
    {\includegraphics[width=0.15\textwidth]{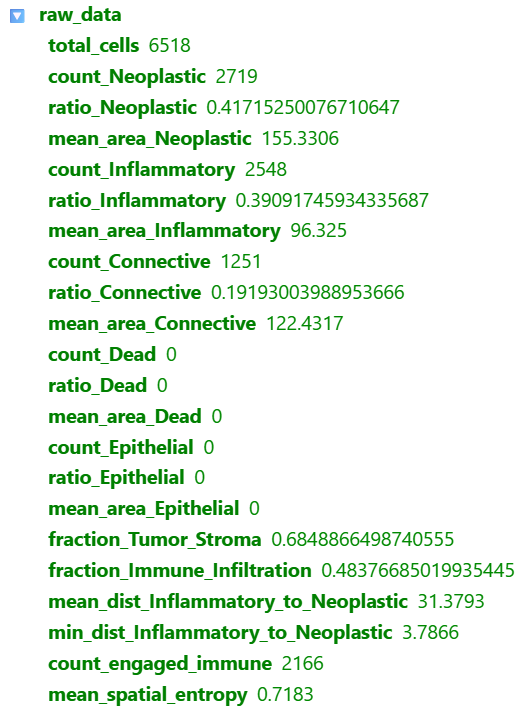}} &
    {\includegraphics[width=0.15\textwidth]{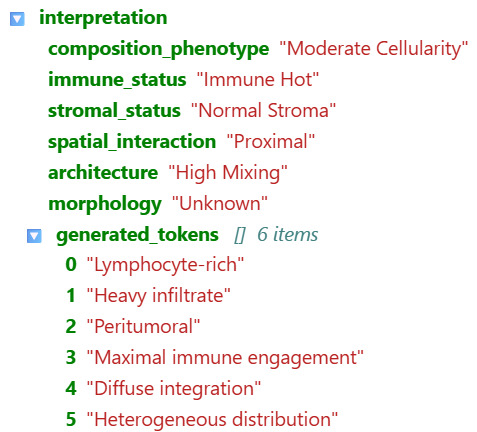}} \\[8pt]
    {\includegraphics[width=0.15\textwidth]{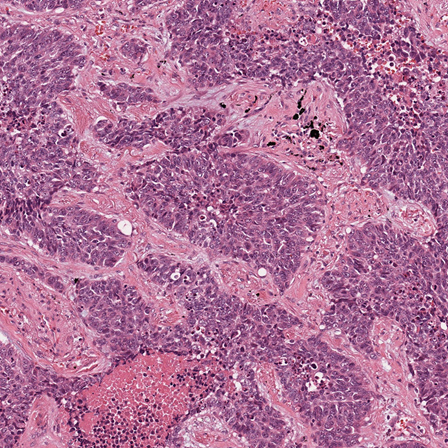}} &
    {\includegraphics[width=0.15\textwidth]{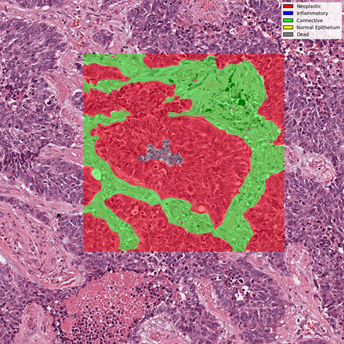}} &
    {\includegraphics[width=0.15\textwidth]{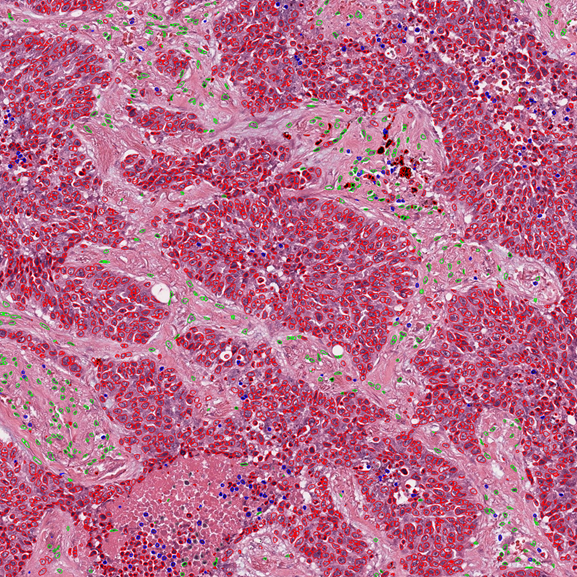}} &
    {\includegraphics[width=0.15\textwidth]{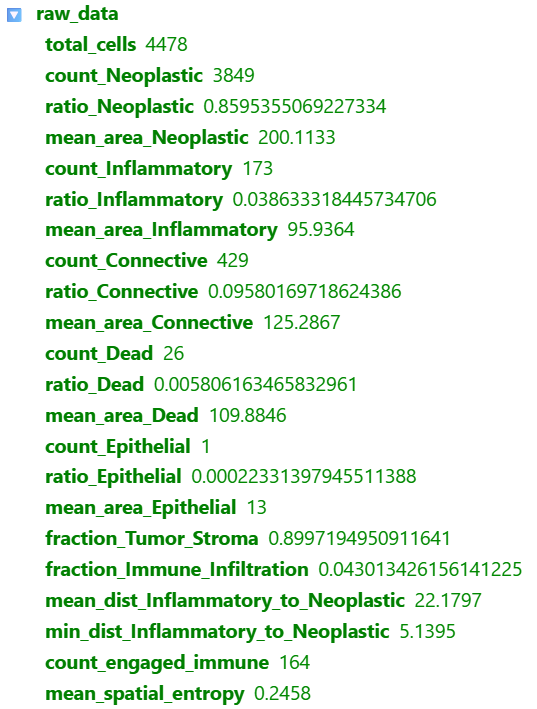}} &
    {\includegraphics[width=0.15\textwidth]{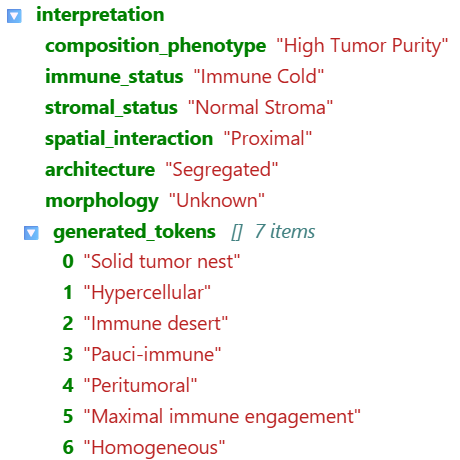}} \\
    {\includegraphics[width=0.15\textwidth]{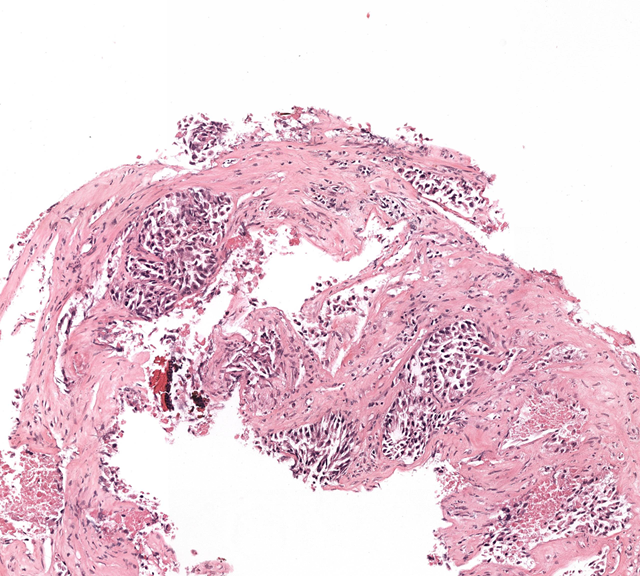}} &
    {\includegraphics[width=0.15\textwidth]{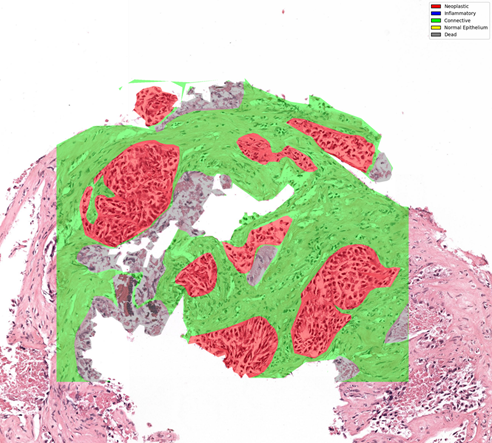}} &
    {\includegraphics[width=0.15\textwidth]{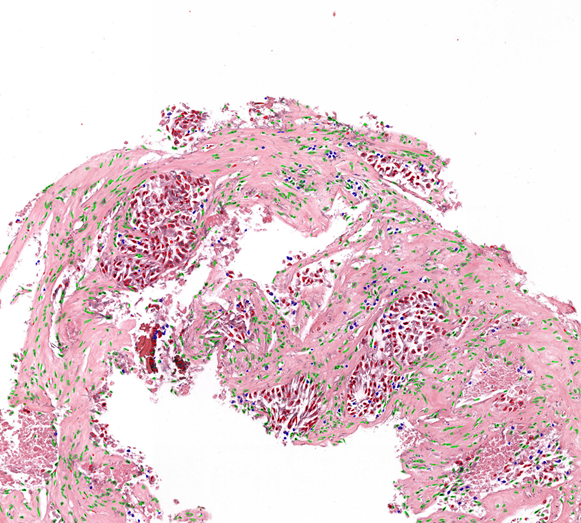}} &
    {\includegraphics[width=0.15\textwidth]{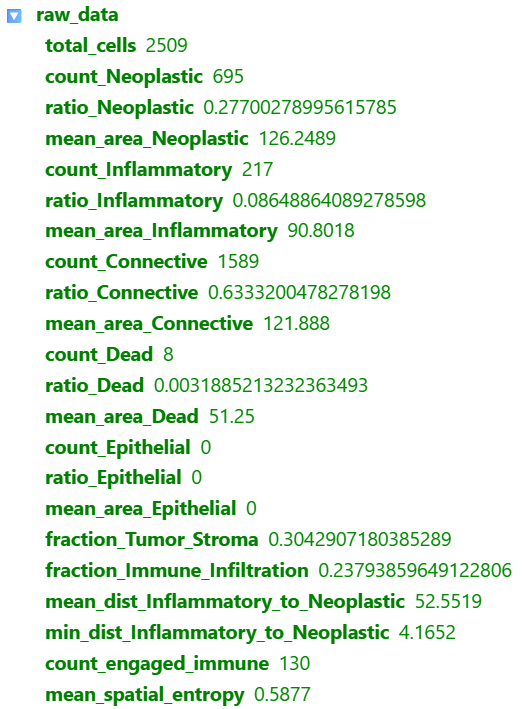}} &
    {\includegraphics[width=0.15\textwidth]{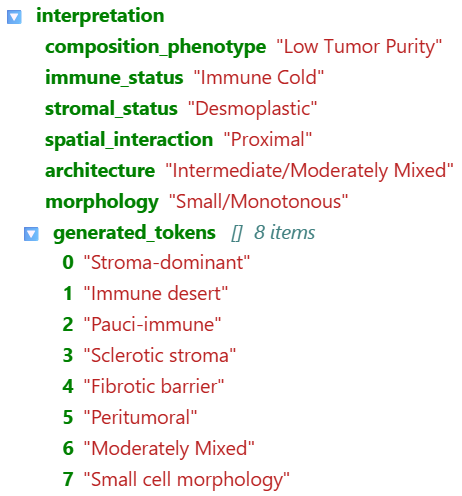}} \\

{\includegraphics[width=0.15\textwidth]{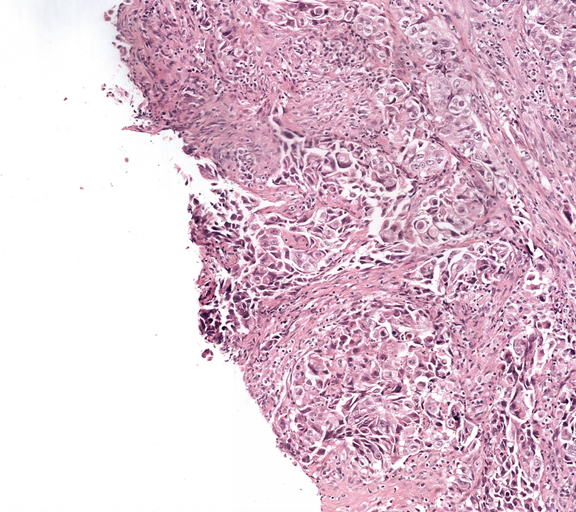}} &
{\includegraphics[width=0.15\textwidth]{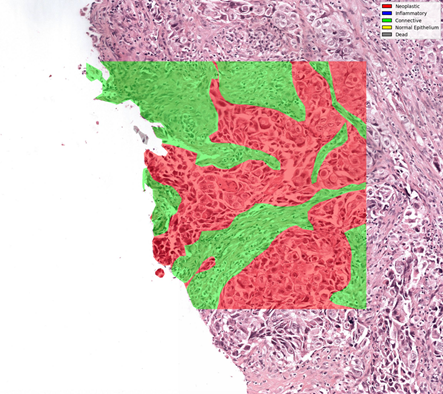}} &
{\includegraphics[width=0.15\textwidth]{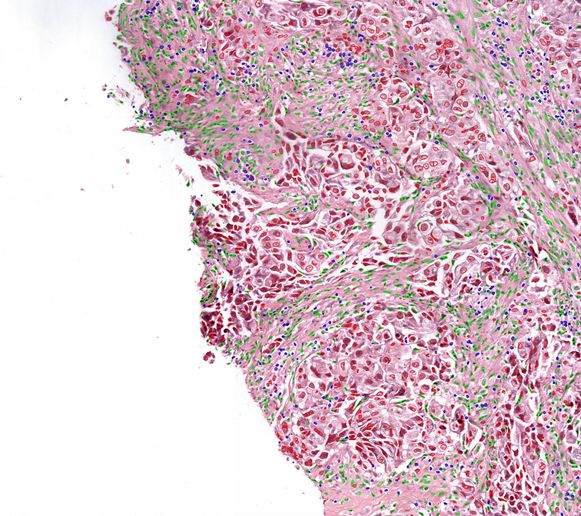}} &
{\includegraphics[width=0.15\textwidth]{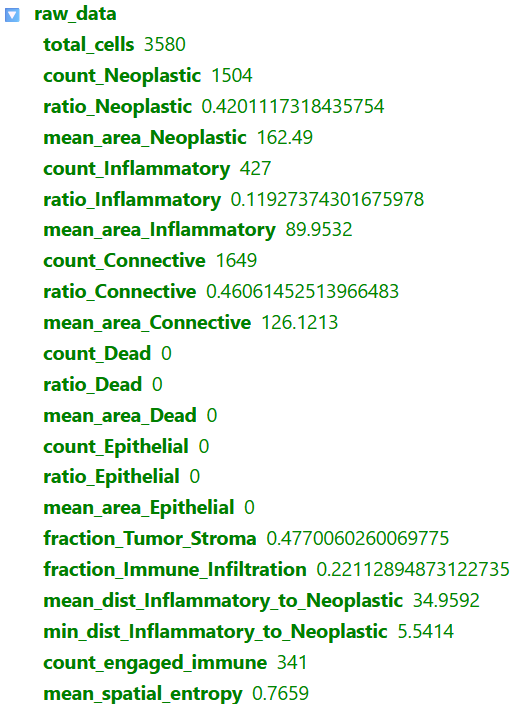}} &
{\includegraphics[width=0.15\textwidth]{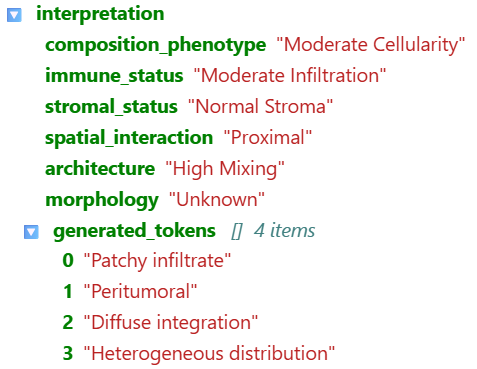}} \\
{\includegraphics[width=0.15\textwidth]{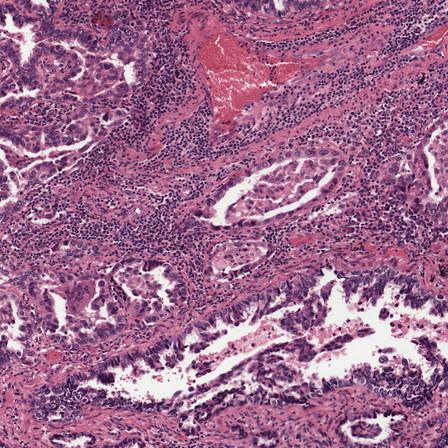}} &
{\includegraphics[width=0.15\textwidth]{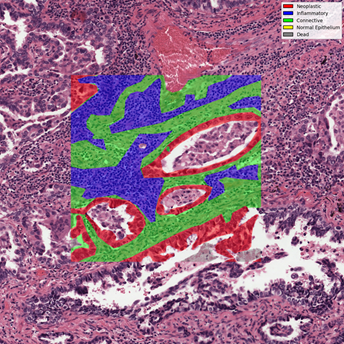}} &
{\includegraphics[width=0.15\textwidth]{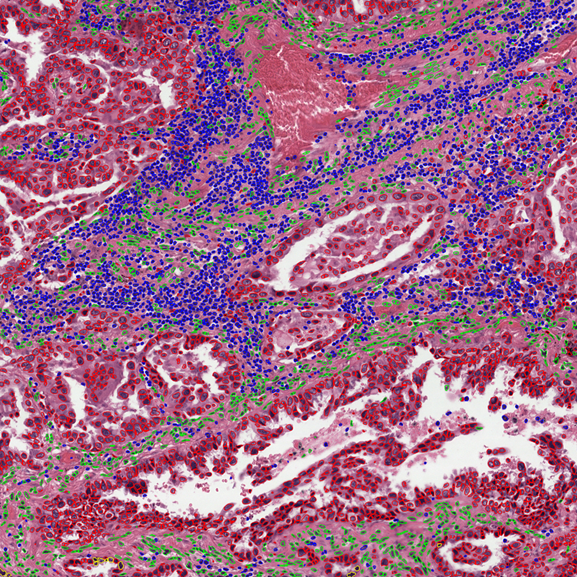}} &
{\includegraphics[width=0.15\textwidth]{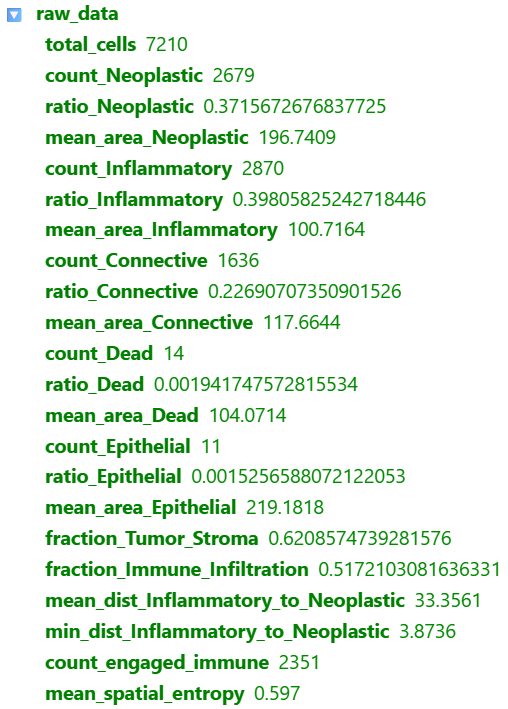}} &
{\includegraphics[width=0.15\textwidth]{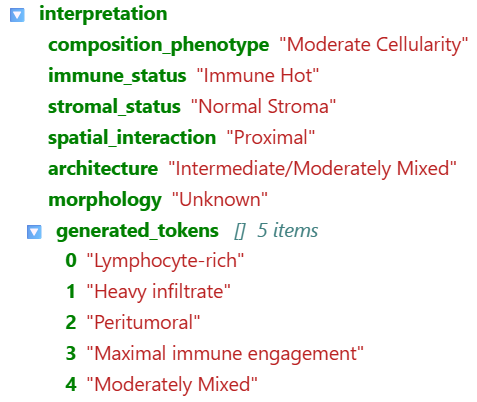}} \\

\end{tabular}
\caption{\textbf{Qualitative TME characterisation on five independent IGNITE NSCLC
tiles (Tiles 1 to 3).}
Each row shows (left to right) the original H\&E tile, ground truth overlay (only provided on areas shown), nucleus outline overlay colour-coded by predicted class (Neoplastic: red; Inflammatory: blue; Connective: green; Normal Epitheleum: yellow; Dead: gray), raw 22-class TME numerical features from \texttt{analyze\_tme\_from
\_maps}, and TME phenotype interpretation and token output generated by
\texttt{translate\_tme\_to\_bio\_tokens} (\Cref{fig:tme_flowchart}).
\textbf{Tile~1 (row 1): 2046$\times$1702 pixels}: ratio of neoplastic (0.42), immune (0.39) and connective (0.19); Immune~Hot, High-Mixing phenotype (entropy 0.72).
\textbf{Tile~2 (row 2): 1792$\times$1792 pixels}: ratio of neoplastic (0.86), immune (0.04) and connective (0.1); High Tumour Purity, Immune~Cold, Segregated architecture (entropy 0.25).
\textbf{Tile~3 (row 3): 2560$\times$2304 pixels}: ratio of neoplastic (0.28), immune (0.09) and connective (0.63); Low Tumour Purity, Immune~Cold, Moderately Mixed architecture (entropy 0.59).
\textbf{Tile~4 (row 4): 2304$\times$2048 pixels}: ratio of neoplastic (0.42), immune (0.12) and connective (0.46); Moderate Cellularity, High Mixing (entropy 0.77).
\textbf{Tile~5 (row 5): 1792$\times$1792 pixels}: ratio of neoplastic (0.37), immune (0.40) and connective (0.23); Immune~Hot, Moderate Cellularity,
Heavy Infiltration, Moderately Mixed architecture (entropy 0.60).
No tile-specific fine-tuning was applied.}
\label{fig:case_study}
\end{figure}

\seg{} processed each patch through the tiled MPP-normalised inference pipeline,
and nuclear instance outlines colour-coded by predicted class were overlaid on
the original H\&E image. \Cref{fig:case_study} presents all five tiles arranged in a four-column layout:
(1) original H\&E patch, (2) nucleus outline overlay colour-coded by predicted class
(Neoplastic: red; Inflammatory: blue; Connective: green; Normal Epitheleum: yellow; Dead: gray), (3) raw 22-class TME numerical features derived from Section \ref{sec:tme} (\texttt{analyze\_tme\_from
\_maps}), and (4) TME phenotype interpretation and token output generated by
\texttt{translate\_tme\_to\_bio\_tokens}.

\FloatBarrier

\FloatBarrier

\begin{table}[htbp]
\centering
\caption{\textbf{BioNeMo-generated TME narratives for the five IGNITE NSCLC tiles
(Tiles 1--5, \Cref{fig:case_study}).} Each narrative is
generated by the LLaMA-3.2-1B model fine-tuned via NVIDIA BioNeMo
(\Cref{sec:bionemo}) from the structured TME feature JSON; raw TME anchor features
shown are the specific numeric values the narrative directly references, allowing
direct spot-checking of each claim against the upstream quantitative output.
A single pathologist reviewed these five narratives as a proof-of-concept
qualitative check.}
\label{tab:bionemo_narratives}
\small
\begin{tabularx}{\textwidth}{@{}
  >{\raggedright\arraybackslash}p{0.8cm}
  >{\raggedright\arraybackslash}p{4.0cm}
  >{\raggedright\arraybackslash}X@{}}
\toprule
Tile & Raw TME Anchor features & Generated narrative \\
\midrule
1 &
Neoplastic ratio 0.42; Connective ratio 0.19; Dist 31.4\,px; Entropy 0.72 &
\textit{``A balanced distribution of neoplastic and non-neoplastic cells is observed (Ratio: 0.42). The region is distinctly 'Immune Hot', situated proximally to the neoplastic margins (Mean distance: 31.4px). Spatially, the tissue shows significant cellular intermixing (Entropy: 0.72).''} \\
\addlinespace
2 &
Neoplastic ratio 0.86; Connective ratio 0.10; Dist 22.2\,px; Entropy 0.25 &
\textit{``The analyzed section is characterized by high tumor purity (Neoplastic ratio: 0.86), forming dense cellular nests. The microenvironment is largely desmoplastic and fibrotic, situated proximally to the neoplastic margins (Mean distance: 22.2px). Inflammatory infiltration is sparse to absent, situated proximally to the neoplastic margins (Mean distance: 22.2px). The spatial architecture is highly compartmentalized and segregated (Entropy: 0.25).''} \\
\addlinespace
3 &
Neoplastic ratio 0.28; Connective ratio 0.63; Dist 52.6\,px; Entropy 0.59 &
\textit{``Low neoplastic cellularity is noted throughout the region (Neoplastic ratio: 0.28). The tumor islands are embedded within a dense, desmoplastic stroma (Connective ratio: 0.63). The region represents an 'Immune Desert', situated proximally to the neoplastic margins (Mean distance: 52.6px). The tissue structure exhibits low spatial diversity (Entropy: 0.59).''} \\
\addlinespace
4 &
Neoplastic ratio 0.42; Connective ratio 0.46; Dist 35.0\,px; Entropy 0.77 &
\textit{``A balanced distribution of neoplastic and non-neoplastic cells is observed (Ratio: 0.42). The microenvironment is highly compartmentalized and homogeneous (Entropy: 0.77). A moderate immune infiltrate is noted, situated proximally to the neoplastic margins (Mean distance: 35.0px). A heterogeneous and mixed spatial distribution is observed (Entropy: 0.77).''} \\
\addlinespace
5 &
Neoplastic ratio 0.37; Connective ratio 0.23; Dist 33.4\,px; Entropy 0.60 &
\textit{``The section shows moderate tumor cellularity (Neoplastic ratio: 0.37). The region is distinctly 'Immune Hot', located in close proximity to the tumor clusters (Dist: 33.4px). The tissue structure exhibits distinct segregation between cell populations (Entropy: 0.60).''} \\
\bottomrule
\end{tabularx}
\end{table}

\FloatBarrier

IGNITE toolkit provides groundtruth annotations only on certain focused area, and did not cover the whole image context. Direct benchmarking is not possible because IGNITE provides tissue level annotations, and the output of \seg{} is nucleus level labels. Visual comparison with IGNITE groundtruth shown an estimation of 80-90\% agreement, with most misclassification on dead or necrotic regions. It is also noted that IGNITE only annotate dense inflammatory regions, where scattered immune cells detected on \seg{} are not annotated. Overall, these five tiles demonstrate that \seg{} produces biologically coherent nucleus outlines
and class assignments across the inter-institutional and inter-scanner stain
variability inherent to the IGNITE toolkit's multi-centric design, without
tile-specific fine-tuning. The consistency of the TME token outputs with the visual
TME phenotype in each tile supports the token robustness argument made in
\Cref{sec:bionemo} because narrative generation is grounded in
population-level statistical features, the
\fw{} pipeline able to maintain clinical accuracy across this public, multi-centric dataset.
The five generated narratives in Table \ref{tab:bionemo_narratives} are validated by a single pathologist as proof of concept. This qualitative evidence also motivates the quantitative 
robustness analyses planned for future work (see Limitations and future work, \Cref{sec:discussion}).

\section{Discussion}
\label{sec:discussion}

\fw{} presents three vital components for characterising TME from routine H\&E histology images: segmentation, TME feature extraction, and narrative generation. Segmentation part addresses the core bottleneck in computational TME analysis: there is far more
TCGA H\&E image data available (1.6M~patches) than there is human-annotated,
pixel-level ground truth to train on (7,901 PanNuke images, 189,744 annotated nuclei).
The three-stage progressive pseudo-label curriculum closes this gap without any
additional manual annotation. Each stage moves to a slightly coarser, broader
domain using pseudo-labels from the model trained in the previous stage. This staged
approach gives the model repeated opportunities to consolidate what it has learned,
and avoids the error accumulation that a single large jump in self-training typically
produces.

UperNet was chosen for segmentation decoder over UNet, SegFormer, and DeepLabV3+ because the resolution shift
from training to inference, PanNuke's fine 0.25\,$\mu$m/pixel up to TCGA-UT's coarser
0.5--1.0\,$\mu$m/pixel, calls for a decoder that can combine features across a wide
range of spatial scales. UperNet's Pyramid Pooling Module and Feature Pyramid Network
do exactly this: they combine fine local detail with broader tissue-level context,
which matters increasingly at coarser resolutions, where nuclei occupy fewer pixels
and their surroundings carry more of the signal. This choice reflects UperNet's
multi-scale design, and testing other head options directly under the same conditions remains an important direction for future evaluation.

An important and practically valuable property of the \fw{} reporting pipeline as a whole
is its robustness to imperfect segmentation. Because the biological tokens and TME narrative
are derived from aggregate statistical features computed over all detected cells in a
patch, the
generated description maintains clinical relevance even when segmentation quality is
reduced by challenging imaging conditions. The IGNITE qualitative validation illustrates
this directly across five tiles spanning three NSCLC histological subtypes and three
contributing scanners: despite inter-institutional and inter-scanner stain variability,
each tile yields a biologically coherent TME phenotype classification and narrative that
correctly reflects the dominant cell composition and spatial organisation observed in
the H\&E image. This
decoupling of narrative quality from single-cell segmentation accuracy is a real
advantage over approaches that generate descriptions directly from raw image pixels:
in those approaches, a local imaging artefact can propagate straight into the output
text with nothing to catch it.

The BioNeMo integration turns quantitative TME output into text a clinical
pathologist can read directly. Because the generated narratives are grounded in
verifiable numerical measurements, rather than image pixels alone, hallucination
risk is substantially reduced, and any narrative claim can be spot-checked against
the feature values in the underlying JSON output, as illustrated for the five IGNITE
NSCLC tiles in \Cref{tab:bionemo_narratives}. The 11-token biological vocabulary also
gives future systems a compact, standardised set of terms to build on, for example in
retrieval-augmented generation, where relevant records are looked up and passed to a
language model alongside the query. \fw{}'s complementary relationship to
image-conditioned copilots such as PathChat, grounding narratives in explicit
quantitative features as opposed to generating text directly from pixels, is discussed
in \Cref{sec:llm_related_work}.



\paragraph{Limitations and future work.}
The current paper presents an architectural framework with preliminary internal
validation across all three components (segmentation, TME feature extraction, and narrative generation). Several important experiments are planned as
follow-up work, grouped below by which part of
the framework they extend: TME characterisation and prognostic validation
(items~1--4), and segmentation refinement (items~5--7).

\textbf{TME characterisation and prognostic validation:}
\begin{enumerate}[leftmargin=*, noitemsep, topsep=2pt]
  \item \textit{Cross-cancer-type validation of the token-translation thresholds:} the
        rule-based thresholds in \Cref{tab:tokens} (e.g.\ the cut-offs defining
        ``Immune Hot'' or ``Moderate Infiltration'') were calibrated against the cancer
        types studied here; whether they generalise across the full range of cancer
        types, or require per-cancer-type recalibration, is an open
        empirical question and the most immediate priority for extending this work.
  \item \textit{Whole-slide aggregation from large-tile narratives:} the
        narrative-generation prompt already reserves metadata fields
        (\texttt{COORDS}, \texttt{MARGIN}, \texttt{OUTCOME}, \Cref{sec:bionemo}) for
        slide coordinates, margin status, and treatment outcome; populating them and
        implementing the aggregation step that combines tile-level narratives into a
        single slide-level description is planned as a concrete next step for prognostication analysis.
  \item \textit{Biological downstream application:} TME features from \seg{} will be
        linked to clinical outcomes (Cox regression, log-rank test) 
        to test whether the generated tokens and narratives carry prognostic value.
  \item \textit{BioNeMo narrative quality:} BLEU-4, ROUGE-L, Cohen's $\kappa$ (two
        independent pathologists), and hallucination rate will quantify narrative
        fidelity, extending the single-pathologist proof-of-concept check in
        \Cref{sec:case_study} into a quantitative evaluation. 
\end{enumerate}

\textbf{Segmentation refinement:}
\begin{enumerate}[resume, leftmargin=*, noitemsep, topsep=2pt]
  \item \textit{Further segmentation validation:} \seg{} has not yet been benchmarked
        against established methods (HoVer-Net, CellPose, StarDist) on public
        instance-segmentation datasets, the HV auxiliary head's individual
        contribution has not been isolated via ablation, and all current results are
        single-run point estimates. Benchmarking on MoNuSeg,
        CryoNuSeg, and CoNSeP (DSC, AJI, PQ), a controlled HV-head ablation, and
        multi-seed statistical reporting are planned together as one more rigorous
        validation pass on the segmentation backbone.
  \item \textit{Pseudo-label quality against human ground truth:} a pathologist-annotated
        sample of 200 TCGA-UT patches will provide human reference labels for direct
        accuracy assessment.
  \item \textit{Scale-aware retraining for the six-scale checkpoint:}
        $\mathcal{M}_3$ currently under-performs $\mathcal{M}_2$ on PanNuke ground
        truth, traced to a single MPP normalisation factor applied uniformly across
        all six TCGA-UT scales during Stage~3 pseudo-label generation. Scale-specific
        normalisation, resolution conditioning, and progressive scale introduction are
        identified as potential fixes. This would extend validated deployment to all six scales, but
        is not a prerequisite for items~1--4. $\mathcal{M}_2$ already supports
        efficient large-tile inference on its own, which is why it is used throughout
        this paper's demonstration (\Cref{sec:case_study}).
\end{enumerate}

Given these limitations and the improvements they motivate, we note that even though $\mathcal{M}_2$ already characterises the TME accurately in practice, a direct accuracy assessment of the pseudo-labels themselves remains important, to give other researchers confidence in using them to train their own models. The quality of BioNeMo-generated narratives is directly tied to the quality of the tokens they are generated from. Since the token-translation thresholds have been qualitatively demonstrated on the IGNITE dataset, the next immediate work is the biological downstream application for IGNITE or other NSCLC datasets, relating these to prognosis, with or without whole-slide aggregation. The potential of \fw{} is extensive, and we have made all the data publicly available to facilitate discovery and advance understanding of TME characteristics from H\&E histopathology images, and their relationship, direct or indirect, with cancer progression and/or patient treatment outcome.

\section{Conclusion}

We presented \fw{}, a unified framework for H\&E-based TME characterisation. Its
central contribution is translating quantitative, segmentation-derived TME features
into an interpretable biological-token vocabulary and clinically grounded narrative
text. \seg{} (a dual-head UNI2-UperNet multiscale segmentation model, trained via a three-stage
progressive pseudo-label curriculum scaling PanNuke supervision to 1.6M
real-world TCGA-UT patches) supplies the per-nucleus class labels this
interpretability layer builds on. It operates over large, variable-size tiles
(\Cref{sec:inference}) (not small fixed patches), so that both the
segmentation backbone and the downstream TME pipeline see the spatial context TME
characterisation requires. A LLaMA-3.2-1B model, fine-tuned via NVIDIA BioNeMo, then
generates the resulting narratives. \fw{} produces biologically coherent phenotype
classifications and narratives despite inter-institutional stain variability and
imperfect segmentation. This robustness comes from grounding narratives in
aggregate population-level statistics,
and it is the central empirical argument for why the token and narrative layer, not
segmentation accuracy itself, is this paper's main contribution.

Internal quantitative validation (\Cref{sec:results}, \Cref{tab:seg_results}) is limited to
the segmentation backbone. All three curriculum-stage checkpoints are evaluated
against PanNuke ground truth and TCGA-UT pseudo-label self-consistency; this
establishes $\mathcal{M}_2$ as the more reliable of the two TCGA-UT-adapted
checkpoints, while $\mathcal{M}_3$ remains limited by a scale-normalisation mismatch, to be resolved with scale-aware retraining. The end-to-end pipeline, spanning feature extraction, token translation,
and narrative generation, is validated qualitatively, on five multi-centric
IGNITE NSCLC tiles. $\mathcal{M}_2$ is used for this demonstration because informal inspection suggested it has more accurate segmentation than $\mathcal{M}_1$ on
external data (visual comparison). $\mathcal{M}_3$ is not used yet because of the diagnosed magnification mismatch described above. The
purpose of this demonstration is to check whether the generated tokens and
narratives are coherent with the observed TME phenotype, not to validate
segmentation accuracy itself, since no ground truth exists for IGNITE to measure
that against. Formal quantitative evaluation of narrative quality, and extending
validated segmentation performance to the full six-scale corpus, both remain open
directions for the companion paper.
The pseudo-labelled TCGA-UT dataset and the \seg{} checkpoints are publicly
released. This is the largest nuclei-level pseudo-labelled TCGA corpus to date, and
it gives other researchers an open, reproducible platform for AI-assisted TME
profiling. The rule-based token-translation logic (\Cref{tab:tokens}), the mechanism
most directly responsible for \fw{}'s interpretability, is fully specified in this
manuscript and reproducible without any additional data. The fine‑tuned BioNeMo narrative checkpoint and prompt templates are planned for release alongside both the formal narrative‑quality evaluation, comprehensive external validation, and their application in biological contexts.

\section*{Ethics Statement}
All TCGA whole-slide images used in this work are publicly available through
the NCI Genomic Data Commons (\url{https://portal.gdc.cancer.gov/}) under
controlled access. The PanNuke dataset is publicly available and uses only
de-identified tissue sections without individual patient identifiers.
The five non-small cell lung cancer (NSCLC) tissue tiles used in the qualitative
validation (Section~\ref{sec:case_study}) are sourced from the publicly available
IGNITE data toolkit~\cite{lucassen2025ignite}, which received its own institutional
ethical approval for public release as documented in the original publication; no
additional institutional approval was required for the secondary use of this
de-identified, publicly released dataset in the present study.
No direct patient contact or prospective clinical intervention was involved.

\section*{Data Availability}
TCGA data: \url{https://portal.gdc.cancer.gov/}.
PanNuke: \url{https://warwick.ac.uk/fac/cross_fac/tia/data/pannuke}.
\bb{} weights: \url{https://huggingface.co/MahmoodLab/uni2-h}.
Pseudo-labelled TCGA-UT dataset:
\url{https://huggingface.co/datasets/mizjaggy18/tcga-ut-cell-instance-semantic}.
\seg{} model weights (three checkpoints):
\url{https://huggingface.co/mizjaggy18/SegTME-UNI2-UperHoVer_PanNuke} (Stage~1),
\url{https://huggingface.co/mizjaggy18/SegTME-UNI2-UperHoVer_TCGA-UT-0} (Stage~2),
\url{https://huggingface.co/mizjaggy18/SegTME-UNI2-UperHoVer_TCGA-UT-012345} (Stage~3).
Model architecture package (including \texttt{UNI2UperHoVer} and \texttt{LargeTilePredictor}, \Cref{sec:inference}): \url{https://pypi.org/project/segtme-uni2/} (\texttt{pip install segtme-uni2}).
The TME feature-extraction and rule-based token-translation logic (\Cref{tab:tokens},
\Cref{eq:ratios,eq:spatial_entropy}) is fully specified in this manuscript and requires no additional data to
reproduce. The BioNeMo-fine-tuned narrative-generation checkpoint and the prompt templates
used for supervised fine-tuning (\Cref{sec:bionemo}) are planned for public release alongside
the companion paper's formal narrative-quality evaluation.

\section*{Declaration of Competing Interests}
The authors declare that they have no known competing financial interests or
personal relationships that could have appeared to influence the work
reported in this paper.

\section*{Declaration of Generative AI and AI-Assisted Technologies
in the Manuscript Preparation Process}
During the preparation of this work the authors used Claude~AI in order to assist with LaTeX code generation, formatting, text editing and flowchart generation based on training code.
After using this tool, the authors reviewed and edited the content including all citations as needed
and take full responsibility for the content of the published article.

\section*{Funding}
This research was supported by GPU computing resources provided through the
\textbf{NVIDIA Academic Grant Programme}. The authors gratefully
acknowledge NVIDIA Corporation for the provision of NVIDIA 8$\times$A100 GPU credits
(Stages~1 and~2 used all eight GPUs; Stage~3 used six of the eight available GPUs
due to cluster scheduling constraints).

\section*{Acknowledgements}
The authors thank the TCGA Research Network for making whole-slide image
data publicly available and the developers of the PanNuke, UNI2-h, and
TCGA-UT datasets for enabling large-scale pathology research.
The authors also gratefully acknowledge the creators of the IGNITE data toolkit
for making their multi-centric NSCLC histopathology dataset publicly available,
enabling the qualitative validation study presented in this work.

\printcredits

\bibliographystyle{elsarticle-num}
\bibliography{segtmeuni2}

@article{graham2019hover,
  title={HoVer-Net: Simultaneous segmentation and classification of nuclei in multi-tissue histology images},
  author={Graham, Simon and Vu, Quoc Dang and Raza, Shan E. Ahmed and Azam, Ayesha and Tsang, Yee Wah and Kwak, Jin Tae and Rajpoot, Nasir},
  journal={Medical Image Analysis},
  volume={58},
  pages={101563},
  year={2019},
  doi={10.1016/j.media.2019.101563}
}

@inproceedings{xiao2018unified,
  title={Unified perceptual parsing for scene understanding},
  author={Xiao, Tete and Liu, Yingcheng and Zhou, Bolei and Jiang, Yuning and Sun, Jian},
  booktitle={Proceedings of the European Conference on Computer Vision (ECCV)},
  pages={418--434},
  year={2018},
  doi={10.1007/978-3-030-01246-5_26}
}

@article{chen2024uni,
  title={Towards a general-purpose foundation model for computational pathology},
  author={Chen, Richard J. and Ding, Tong and Lu, Ming Y. and Williamson, Drew F. K. and Jaume, Guillaume and Song, Andrew H. and Chen, Bowen and Zhang, Andrew and Shao, Daniel and Schuffler, Peter J. and Mahmood, Faisal},
  journal={Nature Medicine},
  volume={30},
  pages={850--862},
  year={2024},
  doi={10.1038/s41591-024-02857-3}
}

@article{chen2024uni2,
  title={UNI2: Towards a universal whole-slide foundation model for pathology},
  author={Chen, Richard J. and Lu, Ming Y. and Ding, Tong and Williamson, Drew F. K. and Jaume, Guillaume and Chen, Bowen and Mahmood, Faisal},
  journal={arXiv preprint arXiv:2406.01647},
  year={2024},
  doi={10.48550/arXiv.2406.01647}
}

@article{gamper2020pannuke,
  title={PanNuke Dataset Extension, Insights and Baselines},
  author={Gamper, Jevgenij and Koohbanani, Navid Alemi and Benes, Ksenija and Graham, Simon and Jahanifar, Mostafa and Khurram, Syed Ali and Azam, Ayesha and Hewitt, Katherine and Rajpoot, Nasir},
  journal={arXiv preprint arXiv:2003.10778},
  year={2020},
  doi={10.48550/arXiv.2003.10778}
}

@article{stringer2021cellpose,
  title={Cellpose: a generalist algorithm for cellular segmentation},
  author={Stringer, Carsen and Wang, Tim and Michaelos, Michalis and Pachitariu, Marius},
  journal={Nature Methods},
  volume={18},
  number={1},
  pages={100--106},
  year={2021},
  doi={10.1038/s41592-020-01018-x}
}

@inproceedings{schmidt2018cell,
  title={Cell detection with star-convex polygons},
  author={Schmidt, Uwe and Weigert, Martin and Broaddus, Coleman and Myers, Gene},
  booktitle={Medical Image Computing and Computer Assisted Intervention (MICCAI)},
  pages={265--273},
  year={2018},
  doi={10.1007/978-3-030-00934-2_30}
}

@article{saltz2018spatial,
  title={Spatial organization and molecular correlation of tumor-infiltrating lymphocytes using deep learning on pathology images},
  author={Saltz, Joel and Gupta, Rajarsi and Hou, Le and Kurc, Tahsin and Singh, Pankaj and Nguyen, Vu and Samaras, Dimitris and Shroyer, Kenneth R. and Zhao, Tianhao and Batiste, Rebecca and Van Arnam, John and {The Cancer Genome Atlas Research Network} and Shmulevich, Ilya and Rao, Arvind U. K. and Lazar, Alexander J. and Sharma, Ashish and Thorsson, V{\'e}steinn},
  journal={Cell Reports},
  volume={23},
  number={1},
  pages={181--193},
  year={2018},
  doi={10.1016/j.celrep.2018.03.086}
}

@article{kather2019deep,
  title={Deep learning can predict microsatellite instability directly from histology in gastrointestinal cancer},
  author={Kather, Jakob Nikolas and Pearson, Alexander T. and Halama, Niels and J{\"a}ger, Dirk and Krause, Jeremias and Loosen, Sven H. and Marx, Alexander and Boor, Peter and Tacke, Frank and Neumann, Ulf Peter and Grabsch, Heike I. and Yoshikawa, Takaki and Brenner, Hermann and Chang-Claude, Jenny and Hoffmeister, Michael and Trautwein, Christian and Luedde, Tom},
  journal={Nature Medicine},
  volume={25},
  number={7},
  pages={1054--1056},
  year={2019},
  doi={10.1038/s41591-019-0462-y}
}

@article{lu2023towards,
  title={A visual-language foundation model for computational pathology},
  author={Lu, Ming Y. and Chen, Bowen and Williamson, Drew F. K. and Chen, Richard J. and Liang, Ivy and Ding, Tong and Jaume, Guillaume and Odia, Igor and Zhang, Andrew and Le, Long P. and Gerber, Georg K. and Mahmood, Faisal},
  journal={Nature Medicine},
  volume={30},
  number={3},
  pages={863--874},
  year={2024},
  doi={10.1038/s41591-024-02856-4}
}

@article{xu2024provgigapath,
  title={A whole-slide foundation model for digital pathology from real-world data},
  author={Xu, Hanwen and Usuyama, Naoto and Bagga, Jaspreet and Zhang, Sheng and Rao, Rajesh and Tristan, Naumann and Wong, Cliff and Gero, Zelalem and Javier, Gonzalez and Poon, Hoifung},
  journal={Nature},
  volume={630},
  pages={181--188},
  year={2024},
  doi={10.1038/s41586-024-07441-w}
}

@inproceedings{bai2017semi,
  title={Semi-supervised learning for network-based cardiac MR image segmentation},
  author={Bai, Wenjia and Oktay, Ozan and Sinclair, Matthew and Suzuki, Hideaki and Rajchl, Martin and Tarroni, Giacomo and Glocker, Ben and King, Andrew and Matthews, Paul M. and Rueckert, Daniel},
  booktitle={Medical Image Computing and Computer Assisted Intervention (MICCAI)},
  pages={253--260},
  year={2017},
  doi={10.1007/978-3-319-66185-8_29}
}

@inproceedings{tarvainen2017mean,
  title={Mean teachers are better role models: Weight-averaged consistency targets improve semi-supervised deep learning results},
  author={Tarvainen, Antti and Valpola, Harri},
  booktitle={Advances in Neural Information Processing Systems (NeurIPS)},
  volume={30},
  year={2017}
}

@inproceedings{grandvalet2004semi,
  title={Semi-supervised learning by entropy minimization},
  author={Grandvalet, Yves and Bengio, Yoshua},
  booktitle={Advances in Neural Information Processing Systems (NeurIPS)},
  volume={17},
  year={2004}
}

@misc{nvidia2023bionemo,
  title={BioNeMo: Large language model framework for life sciences},
  author={{NVIDIA Corporation}},
  howpublished={\url{https://www.nvidia.com/en-us/clara/bionemo/}},
  year={2023}
}

@inproceedings{wolf2020transformers,
  title={Transformers: State-of-the-Art Natural Language Processing},
  author={Wolf, Thomas and Debut, Lysandre and Sanh, Victor and Chaumond, Julien and Delangue, Clement and Moi, Anthony and Cistac, Pierric and Rault, Tim and Louf, R{\'e}mi and Funtowicz, Morgan and Davison, Joe and Shleifer, Sam and von Platen, Patrick and Ma, Clara and Jernite, Yacine and Plu, Julien and Xu, Canwen and Le Scao, Teven and Gugger, Sylvain and Drame, Mariama and Lhoest, Quentin and Rush, Alexander},
  booktitle={Proceedings of the 2020 Conference on Empirical Methods in Natural Language Processing: System Demonstrations (EMNLP)},
  pages={38--45},
  year={2020},
  doi={10.18653/v1/2020.emnlp-demos.6}
}

@inproceedings{ronneberger2015unet,
  title={U-Net: Convolutional Networks for Biomedical Image Segmentation},
  author={Ronneberger, Olaf and Fischer, Philipp and Brox, Thomas},
  booktitle={Medical Image Computing and Computer-Assisted Intervention (MICCAI)},
  pages={234--241},
  year={2015},
  doi={10.1007/978-3-319-24574-4_28}
}

@inproceedings{xie2021segformer,
  title={SegFormer: Simple and Efficient Design for Semantic Segmentation with Transformers},
  author={Xie, Enze and Wang, Wenhai and Yu, Zhiding and Anandkumar, Anima and Alvarez, Jose M. and Luo, Ping},
  booktitle={Advances in Neural Information Processing Systems (NeurIPS)},
  volume={34},
  pages={12077--12090},
  year={2021}
}

@article{huang2023visual,
  title={A visual-language foundation model for pathology image analysis using medical Twitter},
  author={Huang, Zhi and Bianchi, Federico and Yuksekgonul, Mert and Montine, Thomas J. and Zou, James},
  journal={Nature Medicine},
  volume={29},
  pages={2307--2316},
  year={2023},
  doi={10.1038/s41591-023-02504-3}
}

@misc{tcgaUT2024,
  title={TCGA-UT Cell Instance and Semantic Pseudo-label Dataset},
  author={Wan Ahmad, Wan Siti Halimatul Munirah},
  howpublished={HuggingFace Datasets. \url{https://huggingface.co/datasets/mizjaggy18/tcga-ut-cell-instance-semantic}},
  year={2024}
}

@article{komura2022universal,
  author  = {Komura, Daisuke and Kawabe, Akihiro and Fukuta, Keisuke and Sano, Kunio
             and Umezaki, Tatsuya and Koda, Hiroto and Suzuki, Ryohei and Yagi, Yuki
             and Naitoh, Iruru and Minamiguchi, Sachiko and Haga, Hironori
             and Ishikawa, Shumpei},
  title   = {Universal encoding of pan-cancer histology by deep texture representations},
  journal = {Cell Reports},
  year    = {2022},
  volume  = {38},
  number  = {9},
  pages   = {110424},
  doi     = {10.1016/j.celrep.2022.110424}
}

@inproceedings{kirillov2019panoptic,
  author    = {Kirillov, Alexander and He, Kaiming and Girshick, Ross and
               Rother, Carsten and Doll{\'{a}}r, Piotr},
  title     = {Panoptic Segmentation},
  booktitle = {Proceedings of the IEEE/CVF Conference on Computer Vision
               and Pattern Recognition (CVPR)},
  year      = {2019},
  pages     = {9404--9413},
  doi       = {10.1109/CVPR.2019.00963}
}

@article{horst2024cellvit,
  author    = {H{\"o}rst, Fabian and Rempe, Moritz and Heine, Lukas and Seibold, Constantin
               and Keyl, Julius and Baldini, Giulia and Ugurel, Selma and Siveke, Jens
               and Bockmayr, Michael and Samek, Wojciech and Fuchs, Tilman J. and
               Kleesiek, Jens},
  title     = {{CellViT}: Vision Transformers for Precise Cell Segmentation and Classification},
  journal   = {Medical Image Analysis},
  year      = {2024},
  volume    = {94},
  pages     = {103143},
  doi       = {10.1016/j.media.2024.103143}
}

@article{galon2006type,
  author  = {Galon, J{\'e}r{\^o}me and Costes, Anne and Sanchez-Cabo, Fatima and
             Kirilovsky, Amos and Mlecnik, Bernhard and Lagorce-Page{\`e}s,
             Christine and Tosolini, Marie and Camus, Matthieu and Berger, Anne and
             Wind, Philippe and Zinzindohou{\'e}, Franck and Bruneval, Patrick and
             Cugnenc, Paul-Henri and Trajanoski, Zlatko and Fridman, Wolf-Herman and
             Pag{\`e}s, Franck},
  title   = {Type, density, and location of immune cells within human colorectal
             tumors predict clinical outcome},
  journal = {Science},
  volume  = {313},
  number  = {5795},
  pages   = {1960--1964},
  year    = {2006},
  doi     = {10.1126/science.1129139}
}

@article{galon2019approaches,
  author  = {Galon, J{\'e}r{\^o}me and Bruni, Daniela},
  title   = {Approaches to treat immune hot, altered and cold tumours with
             combination immunotherapies},
  journal = {Nature Reviews Drug Discovery},
  volume  = {18},
  pages   = {197--218},
  year    = {2019},
  doi     = {10.1038/s41573-018-0007-y}
}

@article{bengtsson2026digital,
  author  = {Bengtsson, Axel and Andersson, Roland and Andersson, Bodil and
             Ansari, Daniel},
  title   = {Digital quantification of stroma percentage enhances prognostic
             stratification in pancreatic cancer},
  journal = {Surgery in Practice and Science},
  year    = {2026},
  doi     = {10.1016/j.sopen.2026.01.002}
}

@article{keren2018structured,
  author  = {Keren, Leeat and Bosse, Marc and Marquez, Diana and
             Angoshtari, Roshan and Jain, Samir and Varma, Sushama and
             Yang, Soo-Ryum and Kurian, Allison and Van Valen, David and
             West, Robert and Bendall, Sean C. and Angelo, Michael},
  title   = {A Structured Tumor-Immune Microenvironment in Triple Negative
             Breast Cancer Revealed by Multiplexed Ion Beam Imaging},
  journal = {Cell},
  volume  = {174},
  number  = {6},
  pages   = {1373--1387},
  year    = {2018},
  doi     = {10.1016/j.cell.2018.08.039}
}

@article{kronqvist1998morphometric,
  author  = {Kronqvist, P. and Kuopio, T. and Collan, Y.},
  title   = {Morphometric grading of invasive ductal breast cancer.
             I. Thresholds for nuclear grade},
  journal = {British Journal of Cancer},
  volume  = {78},
  number  = {6},
  pages   = {800--805},
  year    = {1998},
  doi     = {10.1038/bjc.1998.581}
}

@article{lu2024pathchat,
  author    = {Lu, Ming Y. and Chen, Bowen and Williamson, Drew F. K. and Chen, Richard J.
               and Liang, Ivy and Ding, Tong and Jaume, Guillaume and Odintsov, Igor
               and Le, Long Phi and Gerber, Georg and Parwani, Anil V. and Zhang, Andrew
               and Mahmood, Faisal},
  title     = {A Multimodal Generative {AI} Copilot for Human Pathology},
  journal   = {Nature},
  year      = {2024},
  volume    = {634},
  pages     = {604--613},
  doi       = {10.1038/s41586-024-07618-3}
}

@article{lucassen2025ignite,
  author    = {Lucassen, Ruben T. and Ciompi, Francesco and Veta, Mitko and Ciompi, Francesco and Bulten, Wouter and Balkenhol, Maschenka and Geessink, Oscar and Smit, Joren and Litjens, Geert and Bejnordi, Babak Ehteshami and Pluim, Josien P. W. and van der Laak, Jeroen and Geijs, Daan J.},
  title     = {A tissue and cell-level annotated {H\&E} and {PD-L1} histopathology image dataset in non-small cell lung cancer},
  journal   = {arXiv preprint arXiv:2507.16855},
  year      = {2025},
  doi       = {10.5281/zenodo.17735903}
}

@inproceedings{Li2022,
  title = {Domain Adaptive Nuclei Instance Segmentation and Classification via Category-Aware Feature Alignment and Pseudo-Labelling},
  ISBN = {9783031164491},
  ISSN = {1611-3349},
  url = {http://dx.doi.org/10.1007/978-3-031-16449-1_68},
  DOI = {10.1007/978-3-031-16449-1_68},
  booktitle = {Medical Image Computing and Computer Assisted Intervention – MICCAI 2022},
  publisher = {Springer Nature Switzerland},
  author = {Li,  Canran and Liu,  Dongnan and Li,  Haoran and Zhang,  Zheng and Lu,  Guangming and Chang,  Xiaojun and Cai,  Weidong},
  year = {2022},
  pages = {715–724}
}

@article{Han2022,
  title = {Multi-layer pseudo-supervision for histopathology tissue semantic segmentation using patch-level classification labels},
  volume = {80},
  ISSN = {1361-8415},
  url = {http://dx.doi.org/10.1016/j.media.2022.102487},
  DOI = {10.1016/j.media.2022.102487},
  journal = {Medical Image Analysis},
  publisher = {Elsevier BV},
  author = {Han,  Chu and Lin,  Jiatai and Mai,  Jinhai and Wang,  Yi and Zhang,  Qingling and Zhao,  Bingchao and Chen,  Xin and Pan,  Xipeng and Shi,  Zhenwei and Xu,  Zeyan and Yao,  Su and Yan,  Lixu and Lin,  Huan and Huang,  Xiaomei and Liang,  Changhong and Han,  Guoqiang and Liu,  Zaiyi},
  year = {2022},
  month = Aug,
  pages = {102487}
}

@inproceedings{Anklin2021,
  title = {Learning Whole-Slide Segmentation from Inexact and Incomplete Labels Using Tissue Graphs},
  ISBN = {9783030871963},
  ISSN = {1611-3349},
  url = {http://dx.doi.org/10.1007/978-3-030-87196-3_59},
  DOI = {10.1007/978-3-030-87196-3_59},
  booktitle = {Medical Image Computing and Computer Assisted Intervention – MICCAI 2021},
  publisher = {Springer International Publishing},
  author = {Anklin,  Valentin and Pati,  Pushpak and Jaume,  Guillaume and Bozorgtabar,  Behzad and Foncubierta-Rodriguez,  Antonio and Thiran,  Jean-Philippe and Sibony,  Mathilde and Gabrani,  Maria and Goksel,  Orcun},
  year = {2021},
  pages = {636–646}
}

\label{LastPage}
\end{document}